\pdfoutput=1

\documentclass[11pt]{article}

\usepackage[final]{acl}
\usepackage{natbib}
\usepackage{times}
\usepackage{latexsym}
\usepackage{proof}
\usepackage{algorithm}
\usepackage[noend]{algorithmic}
\usepackage{amsfonts}
\usepackage{amsmath}
\usepackage{amsthm}
\usepackage{dsfont}

\usepackage{graphicx}
\usepackage{booktabs}
\usepackage{tabularx}
\usepackage{multirow}
\usepackage{longtable}
\usepackage{fontawesome}
\usepackage[T1]{fontenc}
\usepackage{supertabular,booktabs}

\usepackage[utf8]{inputenc}

\usepackage{microtype}

\usepackage{inconsolata}

\usepackage{enumitem}
\usepackage{listings}

\usepackage[utf8]{inputenc}
\def\tcr{\textcolor{red}}
\def\tcb{\textcolor{blue}}
\def\tcg{\textcolor{teal}}
\usepackage{microtype}

\usepackage{listings}
\usepackage{inconsolata}
\usepackage{amssymb}
\usepackage{pifont}
\usepackage{xcolor}
\usepackage{soul}

\setul{0.5ex}{0.3ex} 
\setulcolor{red} 

\newcommand{\tick}[0]{\color{green}{\checkmark}}
\newcommand{\bad}[0]{\color{red}{\ding{55} }}

\lstdefinestyle{promptstyle}{
    basicstyle=\ttfamily\small, 
    breaklines=true,            
    breakatwhitespace=true,     
    columns=flexible,           
    captionpos=b,               
    xleftmargin=0.05\columnwidth,
    xrightmargin=0.05\columnwidth
}

\title{Right for Right Reasons: Large Language Models for Verifiable Commonsense Knowledge Graph Question Answering}

\author{Armin Toroghi\textsuperscript{\textnormal{1}},
  Willis Guo\textsuperscript{\textnormal{1}}, Mohammad Mahdi Abdollah Pour\textsuperscript{\textnormal{1}}, Scott Sanner\textsuperscript{\textnormal{1,2}}\\
  \textsuperscript{1}University of Toronto, Canada\\
  \textsuperscript{2}Vector Institute of Artificial Intelligence, Toronto, Canada\\
  \texttt{\{armin.toroghi, gwillis.guo,} \texttt{m.abdollahpour\}@mail.utoronto.ca}\\
  \texttt{ssanner@mie.utoronto.ca}}

\lstdefinestyle{promptstyle}{
    basicstyle=\ttfamily\small, 
    breaklines=true,            
    breakatwhitespace=true,     
    columns=flexible,           
    captionpos=b,               
    xleftmargin=0.05\columnwidth,
    xrightmargin=0.05\columnwidth
}

\begin{document}
\maketitle
\begin{abstract}
Knowledge Graph Question Answering (KGQA) methods seek to answer Natural Language questions using the relational information stored in Knowledge Graphs (KGs). With the recent advancements of Large Language Models (LLMs) and their remarkable reasoning abilities, there is a growing trend to leverage them for KGQA. However, existing methodologies have only focused on answering factual questions, e.g., \textit{``In which city was Silvio Berlusconi's first wife born?''}, leaving questions involving commonsense reasoning that real-world users may pose more often, e.g., \textit{``Do I need separate visas to see the Venus of Willendorf and attend the Olympics this summer?''} unaddressed. In this work, we first observe that existing LLM-based methods for KGQA struggle with hallucination on such questions, especially on queries targeting long-tail entities (e.g., non-mainstream and recent entities), thus hindering their applicability in real-world applications especially since their reasoning processes are not easily verifiable. In response, we propose Right for Right Reasons ($R^3$), a commonsense KGQA methodology that allows for a verifiable reasoning procedure by axiomatically surfacing intrinsic commonsense knowledge of LLMs and grounding every factual reasoning step on KG triples. Through experimental evaluations across three different tasks—question answering, claim verification, and preference matching—our findings showcase $R^3$ as a superior approach, outperforming existing methodologies and notably reducing instances of hallucination and reasoning errors.
\end{abstract}

\section{Introduction}
Knowledge Graphs (KGs) have been widely used as a structured format for storing and representing relational information. Efficiently querying KGs to obtain the required knowledge is a long-standing problem, for which query languages such as RQL~\cite{karvounarakis2002rql} and  SPARQL~\cite{sparql} have been developed. However, writing queries in these languages requires expertise which limits the accessibility of KGs to inexpert users. Knowledge Graph Question Answering (KGQA)~\cite{zheng2017natural, berant2013semantic, yih2016value} is an established research field that facilitates access to KGs by providing factual answers to natural language (NL) questions using KGs. 

Recently, the promising performance of Large Language Models (LLMs) in reasoning-related tasks has encouraged their application in KGQA research~\cite {kaping, kgr, kbbinder}. While these works have significantly enhanced the performance of KGQA systems, their primary focus has been on addressing factoid questions, such as \textit{"In which city was Silvio Berlusconi's first wife born?"}, which can be answered using only the knowledge graph (KG) facts. However, real-world user queries often extend beyond the factoid knowledge stored in the KG. For example, answering a question such as \textit{"Do I need separate visas to see the Venus of Willendorf and attend the Olympics this summer?"} requires both KG triples indicating the locations of \textit{Venus of Willendorf} and the place where \textit{this summer's Olympics} is taking place, as well as \textit{commonsense reasoning} about how one can identify whether traveling to those countries requires \textit{separate visas} or not. 

Commonsense reasoning is one of the most significant capabilities offered by LLMs~\cite{shen2021generalization, zhao2024large}. Therefore, it may seem straightforward to leverage the LLMs to reason over a set of retrieved KG facts to perform commonsense KGQA. However, LLMs are still susceptible to introducing ungrounded or incorrect information to their reasoning process -- a phenomenon called \textit{hallucination}~\cite{ye2023cognitive, tonmoy2024comprehensive}. In conducting commonsense KGQA, LLMs may exhibit hallucinations both by introducing ungrounded factual information as well as making incorrect commonsense inferences. Hence, verifiability of the reasoning process is crucial to ensure the reliability of the final answer, especially in high-stakes applications. Regrettably, none of the existing LLM-enhanced KGQA methodologies answer queries following a verifiable scheme. 

In this paper, we experimentally show that the performance of existing KGQA methods is critically hindered by the hallucination issue when faced with questions involving commonsense reasoning. This issue is particularly exacerbated for questions about long-tail knowledge, i.e., questions targeting obscure or recent entities, and personalized questions. To address this challenge, we introduce \textit{Right for Right Reasons $(R^3)$}, a verifiable methodology for performing KGQA using LLMs. $R^3$ makes both aspects of commonsense KGQA reasoning, factoid steps and commonsense inferences, verifiable. For the commonsense inference aspect, it axiomatically surfaces the commonsense knowledge required for answering the question that is intrinsic to the LLM parameters. Also, it casts the KGQA task into a tree-structured search in which all factual reasoning steps are enforced to be grounded on a subset of the relevant KG triples which enables the verification of factual reasoning steps. We compare $R^3$ against current LLM-based KGQA methodologies and pure LLM methods on three different tasks: question answering, claim verification, and KG-based preference matching. The results demonstrate that $R^3$ leads to a considerable reduction in hallucination and reasoning errors while often improving accuracy and offering robustness to entity popularity.

\section{Background}

\subsection{Reasoning with Large Language Models}
Despite being originally designed for text generation, LLMs have shown outstanding performance when applied to several other NLP sub-fields~\cite{chang2023survey}. Particularly, the reasoning capability of LLMs has attracted considerable interest in AI research~\cite{arora2022ask, sun2022paradigm, xu2023llms}. Several works have studied different reasoning skills of LLMs such as arithmetic reasoning~\cite{yuan2023well}, logical reasoning~\cite{liu2023evaluating}, and commonsense reasoning~\cite{bian2023chatgpt, shen2023experimental}. These abilities make LLMs apt candidates for being used as a reasoner in 
specialized tasks~\cite{ren2023robots, song2023llm, clusmann2023future}.
\subsection{Commonsense Question Answering}
The general knowledge and conception about the world that humans possess, and their ability to reason about it is called commonsense reasoning and is a crucial cognitive ability of humans. It is also an important reasoning skill based on which AI agents are evaluated~\cite{liu2021generated, bauer22, wang2023car}. LLMs have shown outstanding commonsense reasoning skills and the gap between their performance and humans on available datasets has narrowed substantially~\cite{guan2023multi, bian2023chatgpt}. 
Most of these datasets such as CommonsenseQA~\cite{talmor2018commonsenseqa} and PhysicalQA~\cite{bisk2020piqa} contain questions about concepts rather than entities. Recently, StrategyQA~\cite{geva2021did} and Creak~\cite{onoe2021creak} have been proposed as datasets for commonsense reasoning about entities that can be used to introduce commonsense reasoning to KGQA.

\subsection{Knowledge Graph Query Answering}
Answering questions using the relational information of KGs has recently gained significant attention~\cite{wang2024knowledge, toroghi2024bayesian}, with its applications ranging from healthcare~\cite{guo2022medical} to recommendation~\cite{toroghi2023bayesian}. 
Most existing works on the task of answering NL queries using the KG facts, known as KGQA, focus on converting the NL query into a structured formal query in a language such as SPARQL, executing the query to retrieve the required knowledge, and finally reasoning over the retrieved facts to obtain the final answer. This idea, referred to as semantic parsing~\cite{lan2021survey, gu2022arcaneqa, cheng2022binding}, often involves the data and computationally expensive process of fine-tuning with thousands of labeled examples~\cite{chen2021retrack, shu2022tiara}. Recently, KB-BINDER has suggested a training-free semantic parsing methodology using the in-context learning ability of LLMs with few-shot examples~\cite{li2023few}. Novel LLM-based methods beyond semantic parsing approach have also been proposed. KAPING~\cite{kaping} introduced an efficient LLM-enhanced KGQA model that finds the relevant sub-graph to the query via dense retrieval and uses the LLM to reason over it in a zero-shot manner. KGR~\cite{kgr} proposed the idea of allowing LLMs to make claims, retrofitting those claims on the KG facts, and finally reasoning using the corrected claims. However, all existing works on KGQA are designed to answer factoid queries, and none of them has considered queries involving commonsense reasoning.

\section{Methodology}
\subsection{Problem Formulation}
In this paper, we propose a methodology for performing commonsense KGQA that is easily extended to other related tasks such as KG-based preference matching. The input to the problem is a NL sentence posed by the user that can be either a question in the form of an interrogative sentence, or a claim or need expressed as an imperative sentence. We use the term \textit{query}, denoted by $q$, to refer to the input in all cases. The query mentions a set of anchor entities $\mathcal{E}^q$.
A KG $K=( \mathcal{E}, \mathcal{R})$ is assumed to be given, where $\mathcal{E}$ and $\mathcal{R}$ denote its set of entities and relations respectively, such that $\mathcal{E}^q \subset \mathcal{E}$. The objective is to follow a sequence of reasoning steps $\mathcal{S}^q$ to find $a^q \in \mathcal{O}^q$, the answer to the query, such that verifying the correctness of every $s^{q}_{i} \in \mathcal{S}^q$ is possible. Here, $\mathcal{O}^q$ denotes the set of possible options.

\begin{figure*}
    \centering
    \includegraphics[width=\textwidth]{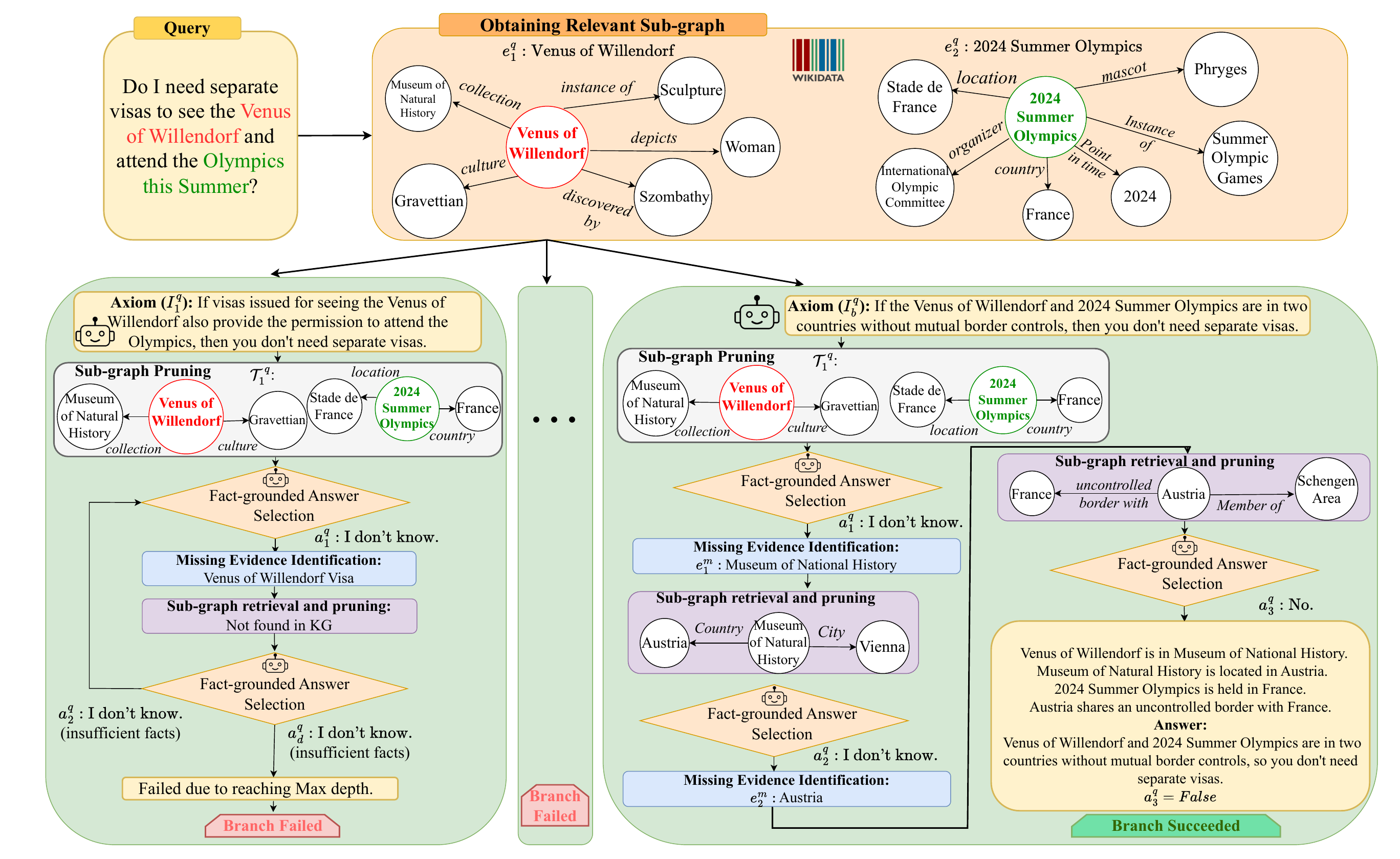}
    \caption{Workflow of commonsense KGQA procedure using $R^3$. After extracting entities from the query and retrieving their relevant sub-graphs, a commonsense axiom is surfaced from the LLM that guides the reasoning branch. After pruning sub-graphs to obtain their relevant facts to the axiom, an iterative process using the LLM is executed to either provide a fact-grounded answer or identify missing information and retrieve it from the KG. If the answer is not found after a certain depth, a new axiom is surfaced to guide a new branch.}
    \label{fig:diagram}
\end{figure*}

\subsection{Right for Right Reasons}
\label{sec:r3}
Our proposed method casts the problem of commonsense KGQA as a tree-structured search, in which every reasoning step is either grounded on KG facts, or based on surfaced commonsense axioms, a key property that makes the reasoning procedure completely \textit{verifiable}. The overall workflow of $R^3$ for answering a query is shown in Figure \ref{fig:diagram}. In brief, $R^3$ first identifies the anchor entities of a query and obtains the relevant sub-graph for these entities. Next, it surfaces a commonsense axiom from the LLM that will guide the reasoning steps in that branch of the search tree. Then, at each depth level of the tree, it checks whether the commonsense axiom can be satisfied with the available KG facts, and if possible, provides an answer grounded on a subset of them. If the available KG triples are insufficient, by backward-chaining from the axiom, it selects the next entity to obtain its relevant KG sub-graph to continue the search. Each branch can continue up to a maximum depth, and if an answer is not obtained at its bottom, a new commonsense axiom will be surfaced which will guide search in a new branch until the search tree reaches its maximum breadth. Components of $R^3$ are explained here, and a series of analyses on their roles and significance are provided in Appendix \ref{sec:appendix-ablation}.
\subsubsection{Obtaining Relevant Sub-graph}
The query answering process begins by extracting $\mathcal{E}^q$ from $q$. Most existing works perform this extraction using entity linking techniques~\cite{ li2020efficient, ayoola2022refined}. However, since existing entity linkers may fail to extract recent or obscure entities from the query, we also use an LLM-based module with few-shot examples to obtain another set of entity names, and consider the union of the two sets as the final set of entities. Formally, 
\begin{equation}
\label{eq:el}
    \mathcal{E}^q = \mathtt{EL}(q, \mathcal{K}) \cup \mathtt{LLM_{E}}(q),
\end{equation}
where $\mathtt{EL}$ is an entity linker module and $\mathtt{LLM_E}$ is the LLM-based module that identifies anchor entities mentioned in $q$. Once the anchor entities are identified, we extract $\mathcal{K}^q \subset \mathcal{K}$, the sub-graph of $\mathcal{K}$ within the 1-hop neighborhood of $\mathcal{E}_q$.
\begin{equation}
\label{eq:subgraph}
\mathcal{K}^q = \{(h, r, t) | (h, r, t) \in \mathcal{K} \land h \in \mathcal{E}^q \}.
\end{equation}

\subsubsection{Surfacing Commonsense Axioms}
The commonsense knowledge that LLMs have obtained during their training process is intrinsic to their parameters, and they can use it to answer queries given a set of retrieved facts. Existing LLM-based methods that are designed for tackling the factoid KGQA problem can approach commonsense KGQA using this intrinsic capability of their LLM component. However, since the set of commonsense axioms the reasoner has used is not known, the reasoning process is not verifiable. To address this issue, $R^3$ axiomatically surfaces this intrinsic knowledge of the reasoner and uses it to guide the reasoning process. In other words, its reasoner is enforced to state the premises required for concluding an answer as a set of atomic factoid clauses and try to find the answer by identifying whether those clauses are satisfied when their variables are grounded on the KG entities, and their predicates and functions on KG relations. For example, when given a query \textit{"Would it make sense for Virginia Raggi to ask for a quinceañera?"}, the reasoner surfaces the axiom: \textit{"If Virginia Raggi is a girl from Latin America and her age is near 15, it would make sense for her to ask for a quinceañera."} 

Formally, given $\mathcal{E}^{q} = \{e_{1}^{q}, ..., e^{q}_{|\mathcal{E}_{q}|}\}$, a commonsense axiom $I_q$ is an NL representation of the First-Order Logic (FOL) expression
\begin{multline}
\left( \bigwedge\limits_{i=1}^{|\mathcal{P}|}\bigwedge\limits_{j=1}^{|\mathcal{E}|}P_i(e_j) \right) \land \left( \bigwedge\limits_{i=1}^{|\mathcal{F}|}\bigwedge\limits_{j=1}^{|\mathcal{E}|}F_i(e_j) 
\, \langle\mathit{op}_j^i\rangle \, e_j^i \right) \\ \implies a_q,
\end{multline}
in which $\mathcal{P} = \{ P_1, ..., P_{|P|} \}$ is the set of predicates, $\mathcal{F} = \{ F_1, ..., F_{|F|} \}$ is the set of functions, $\langle\mathit{op}_j^i\rangle \in \{ =, \neq, <, \leq, >, \geq \}$ is a (dis)equality operator or comparison operator if the function value is numeric, $e_j^i$ is the entity compared to the function evaluation, and $a_q$ is the answer to the query or claim. These relations and functions are all atomic clauses that can be checked against the KG triples. 
\subsubsection{Sub-graph Pruning}
 Once a commonsense axiom is surfaced, $R^3$ tries to identify the satisfiability of the premises based on the KG triples. Since the number of triples in $\mathcal{K}^q$ may be large, we need to first prune the set of available KG triples. To this end, as in ~\cite{kaping}, we use off-the-shelf dense retrievers~\cite{song2020mpnet, karpukhin2020dense, xiong2020approximate} to obtain $\mathcal{T}^{q}_{i} \subset \mathcal{K}^q$, the subset of triples that have the most semantic similarity to the commonsense axiom $I^{q}_{i}$. Since filtering triples by only considering semantic similarity may lead to a high risk of losing some useful triples, we also use an LLM module with few-shot examples to pick relevant triples to the axiom from a subset of the sub-graph triples to reduce the chance of this information loss. Formally we have the Sub-graph Pruning module $\mathtt{SGP}$ as
\begin{equation}\label{equation2}
\begin{split}
    & \mathtt{SGP}({I}^{q}_{i},\mathcal{K}^{q}) = \\ &\;\;\; \mathtt{top\textrm{-}k}_{t \in \mathcal{K}_q} (sim(\boldsymbol{t}, \boldsymbol{I}^{q}_{i})) \cup \mathtt{LLM}_{T}(\mathcal{K}^{q}, {I}^{q}_{i}),\\
    &\mathcal{T}^{q}_{i} = \mathtt{SGP}({I}^{q}_{i},\mathcal{K}^{q}),
    \end{split}
\end{equation}
in which $sim$ denotes the Euclidean similarity between $\boldsymbol{t}$ and $\boldsymbol{I}^{q}_{i}$, the embedding vectors of the triple $t$ and the axiom $I^{q}_{i}$, $\mathtt{top}\textrm{-}k$ operator returns the first $k$ elements of the sorted list of
triples by their similarity score in descending order, and $\mathtt{LLM}_{T}$ is an LLM-based module that returns a subset of $\mathcal{K}^{q}$ that are relevant to $I^{q}_{i}$.
\subsubsection{Fact-Grounded Answer Selection}
After surfacing the commonsense axiom $I^{q}_{i}$, and obtaining the set of relevant triples $\mathcal{T}^{q}_{i}$, $R^3$ tries to identify whether all premises in the axiom can be satisfied by grounding them on the relevant triples, in which case the answer to the query is \textit{"True"}, or at least one of the premises is unsatisfied, making the answer \textit{"False"}. If the axiom is in a disjunctive form, the answer becomes \textit{"True"} as soon as each disjunctive clause is completely satisfied. In all these cases, $R^3$ returns the answer, and the reasoning process is terminated. For multiple-choice queries, the process is repeated for each option until an option satisfies all premises. However, if the satisfiability of any of the premises is not identifiable by the current set of facts, instead of returning a guessed answer that encourages hallucination, the answer will remain undetermined. In this case, the set of current facts is insufficient for grounding all premises, so the reasoning process must continue to the next depth level. Formally, the answer  $a^q \in \{ \text{"True"},\text{"False"},\text{"I don't know"} \}$ is determined by
\begin{equation}
a^q = \mathtt{answer}(q,\boldsymbol{I}^q_{i},\mathcal{T}^{q}_{i}),
\end{equation}
where $\mathtt{answer}$ is the LLM-based module determining the final answer. 

\subsubsection{Missing Evidence Identification}
The set of retrieved facts may be insufficient in two cases: either the query targets a different entity, as in multi-hop questions, or the facts required for grounding at least one premise were mistakenly pruned. In this step, the reasoner is asked to consider the set of unsatisfied premises and the existing facts to first identify what additional evidence must be obtained that is currently missing. Then, it has to identify the anchor entity $e_m$ that its triples can provide the missing information.
If the anchor entity is already in, $\mathcal{E}_q$, the next top $k$ relevant facts about it will be picked for the next step. Otherwise, the reasoner is asked to propose the next entity and extract its name from $\mathcal{K}^q$.
The next entity is then added to $\mathcal{E}_q$, and the process of sub-graph extraction and pruning is executed for it. Formally,
\begin{align}
e^{m} & =  \texttt{MEI}(q,I^q_{i},\mathcal{T}_{i}), \label{eq:MEI} \\
\mathcal{E}^{q}_{j+1}  & = \mathcal{E}^{q}_{j} \cup \{ e^{m}\},\nonumber \\
\mathcal{K}^{q}_{j+1} &= \mathcal{K}^{q}_{j} \cup \{(h,r,t) | (h,r,t) \in \mathcal{K} \land h \in e^m\},\nonumber \\
\mathcal{T}_{i} &= \mathcal{T}_{i} \cup  \mathtt{SGP}(\mathcal{K}^{q}_{j+1}, I^{q}_{i}), \nonumber
\end{align}
where $\texttt{MEI}$ is the module identifying entity $e^m$. This procedure continues until an answer is found or the maximum depth is reached for the branch. In case the maximum depth for a branch is reached without obtaining an answer, a new commonsense axiom will be generated to form a new branch.

\subsection{Comparison to Existing KGQA Methods}
$R^3$ is the first KGQA approach that supports commonsense queries in a verifiable manner, since every factual reasoning step is grounded on particular KG triples, and its commonsense reasoning assumptions are surfaced in the form of axioms. Although KGR~\cite{kgr} retrofits its factual claims on the KG, its commonsense reasoning process is implicit. Semantic parsing methods are only designed for factoid queries and cannot address commonsense queries. Finally, KAPING~\cite{kaping}, despite its strong performance on single-hop factoid queries, cannot answer multi-hop questions because it has no particular mechanism for traversing the KG. A summary of key properties of existing KGQA methods and their comparison to $R^3$ is provided in Table \ref{tab:methods_comparison}. 
\begin{table*}[htbp]
    \centering
    \begin{tabular}{|l|ccccc|}
    \hline
    Method      & Factoid QA& 	Verifiability	& Commonsense	& No training & Multi-hop  \\
    \hline
    Classical Semantic Parsing    & \tick & \bad & \bad & \bad & \tick \\ 
      KB-BINDER   & \tick &\bad &\bad & \tick & \tick\\ 
      KAPING   &  \tick \ & \bad&  \tick \  &  \tick \ & \bad \\ 
      KGR & \tick \ & \bad & \tick & \tick & \tick \\
       $R^3$   &  \tick \ &   \tick \  &  \tick\  &   \tick \ & \tick \\ 
       \hline

    \end{tabular}
    \caption{Comparison of $R^3$ properties against existing KGQA Methods}
    \label{tab:methods_comparison}

\end{table*}

\section{Experiments}
\label{sec:experiments}
We empirically evaluate $R^3$ on three tasks: Question answering, claim verification, and KG-based preference matching. All tasks are closely related to KGQA and involve commonsense reasoning. We release all our implementation codes and data
\footnote{\label{url:repo}\url{https://anonymous.4open.science/r/RRR-4F47/}}.
\subsection{Task Description}
\textbf{Question Answering}. In this task, a question requiring commonsense reasoning formed around some KG entities is asked. The reasoner is required to find the answer, which is either \textit{"Yes"} or \textit{No}.

\vskip 4pt
\noindent
\textbf{Claim Verification} Claim verification is very similar to question answering. Here, an imperative sentence including a claim about some entities is stated. The reasoner has to use the KG facts to decide whether the claim is \textit{"Correct"} or \textit{"Incorrect"}.

\vskip 4pt
\noindent
\textbf{KG-based Preference Matching} 
In this task, a query explaining the user's preference and a personal KG containing evidence about the user's preferences and restrictions is presented to the reasoner. The reasoner has to choose the item that matches both the user's query and their personal restrictions.

\subsection{Datasets}
Due to the lack of existing datasets, we modify three existing datasets to make them suitable for our tasks and make them publicly available to encourage research on commonsense KGQA. Examples of these modifications are shown in Table \ref{dataset}.

\begin{table*}[t!]
\centering
\small
\begin{tabular}{@{}lllcccccccc@{}}
\toprule
\multirow{2}{*}{\textbf{Task}} &
  \multirow{2}{*}{\textbf{Model}} &
  \textbf{} &
  \multicolumn{2}{c}{\textbf{Accuracy}} &
  \textbf{} &
  \multicolumn{2}{c}{\textbf{FActScore}} &
  \multicolumn{1}{l}{} &
  \multicolumn{2}{c}{\textbf{Reasoning}} \\ \cmidrule(lr){4-5} \cmidrule(lr){7-8} \cmidrule(l){10-11} 
 &
   &
   &
  \textbf{Original} &
  \textbf{Long-Tail} &
   &
  \textbf{Original} &
  \textbf{Long-Tail} &
  \multicolumn{1}{l}{} &
  \textbf{Original} &
  \textbf{Long-Tail} \\ \midrule
\multirow{6}{*}{\textbf{\begin{tabular}[c]{@{}l@{}}Question \\ Answering\end{tabular}}} &
  0-shot CoT &
   &
  0.70 &
  0.32 &
   &
  0.63 &
  0.54 &
  \multicolumn{1}{l}{} &
  0.90 &
  0.89 \\
 &
  2-shot CoT &
   &
  0.70 &
  0.43 &
   &
  0.64 &
  0.52 &
  \multicolumn{1}{l}{} &
  0.92 &
  0.90 \\
 &
  KAPING &
   &
  0.72 &
  0.67 &
   &
  0.74 &
  0.59 &
   &
  0.86 &
  0.83 \\
 &
  KB-BINDER &
   &
  0.11 &
  0.08 &
   &
  - &
  - &
   &
  - &
  - \\
 &
  KGR &
   &
  0.39 &
  0.13 &
   &
  0.61 &
  0.47 &
   &
  0.70 &
  0.65 \\
 &
  $R^3$ &
   &
  \textbf{0.82} &
  \textbf{0.73} &
   &
  \textbf{0.97} &
  \textbf{0.96} &
   &
  \textbf{0.97} &
  \textbf{0.95} \\ \midrule
\multirow{6}{*}{\textbf{\begin{tabular}[c]{@{}l@{}}Claim \\ Verification\end{tabular}}} &
  0-shot CoT &
   &
  0.89 &
  0.35 &
   &
  0.76 &
  0.59 &
  \multicolumn{1}{l}{} &
  0.93 &
  0.91 \\
 &
  2-shot CoT &
   &
  \textbf{0.92} &
  0.41 &
   &
  0.78 &
  0.58 &
  \multicolumn{1}{l}{} &
  0.93 &
  0.92 \\
 &
  KAPING &
   &
  0.91 &
  0.81 &
   &
  0.81 &
  0.75 &
   &
  0.90 &
  0.88 \\
 &
  KB-BINDER &
   &
  0.35 &
  0.14 &
   &
  - &
  - &
   &
  - &
  - \\
 &
  KGR &
   &
  0.80 &
  0.20 &
   &
  0.70 &
  0.58 &
   &
  0.74 &
  0.71 \\
 &
  $R^3$ &
   &
  0.85 &
  \textbf{0.85} &
   &
  \textbf{0.98} &
  \textbf{0.98} &
   &
  \textbf{0.97} &
  \textbf{0.96} \\ \bottomrule
\end{tabular}
\caption{Results for all methods on the question answering and claim verification task on both the original and modified (long-tail) queries. FaCTScore and Reasoning are human evaluated metrics.}
\label{qa-cv-results}
\end{table*}
\vskip 4pt
\noindent
\textbf{Question Answering}
Early KGQA datasets consisted of simple questions that can be answered using a single KG triple. Recently, datasets containing more complex questions by introducing multi-hop reasoning have been proposed~\cite{webquestions, lcquad, grailqa}. However, all KGQA datasets contain factoid questions, which do not require commonsense reasoning to answer~\cite{guo2024cr}. Some datasets exclusively focus on evaluating commonsense reasoning~\cite{talmor2018commonsenseqa, boratko2020protoqa, sap2019socialiqa}, but their questions target concepts (e.g., river, mountain, etc.) rather than KG entities (e.g., specific people, locations, etc.).

To overcome this challenge we modify StrategyQA~\cite{strategyqa}, a QA dataset with Yes/No questions that target entities from Wikipedia{\footnote{\url{https://www.wikipedia.org/}}} articles. We select a subset of 150 questions for which the required factual knowledge for answering them is present in Wikidata{\footnote{\url{https://www.wikidata.org/}}} or that can be rewritten as such queries by targeting them on new entities. The questions mostly target famous entities that LLMs can answer using their internal knowledge without hallucinating or even needing a KG. Since we are particularly interested in studying the hallucination behavior of LLM-based KGQA methods on long-tail knowledge, for each query, we also write a counterpart targeting long-tail knowledge by substituting its entities with obscure entities of the same types. We use the number of Wikidata triples and Google Search results as measures of popularity.

\vskip 4pt
\noindent
\textbf{Claim Verification}
For KG-based claim verification, we use Creak~\cite{creak}, a dataset containing True/False claims written by crowd workers using Wikipedia. We follow a similar procedure applied to the QA dataset to select 150 claims and write their long-tail counterparts.

\noindent
\textbf{KG-based Preference Matching}
Recipe-MPR~\cite{zhang2023recipe} is a preference matching dataset that contains NL queries expressing a user's preference toward recipes and often targeting multiple aspects. The reasoner has to choose the recipe that satisfies all aspects among five options. The multi-aspect nature of its queries and the necessity for performing logical reasoning make it a relevant dataset to our work. However, its queries are not personalized, meaning that the correct recipe does not require reasoning over the user's preferences and restrictions beyond those stated in the query. In real-world applications, the \textit{"correct"} item is different for each user considering their personal preferences and restrictions. To bridge this gap, we first extract 100 queries from Recipe-MPR dataset that require commonsense reasoning and add a synthetic personal KG for the user posing the query. We also add a sixth option that matches every preference aspect of the query but violates at least one personal requirement that can be inferred from the user's personal KG. 

\begin{table*}[t!]

\small
\centering
\begin{tabular}{|p{4cm}|p{11cm}|}
\hline
\multicolumn{2}{|c|}{\textbf{Question Answering}} \\ 
\hline

 \textbf{Exemplar Original Question} & Did \tcb{Alan Turing} suffer from the same fate as \tcb{Abraham Lincoln}?\\ 
 \hline
 \textbf{Exemplar Modified Question}&  Did \tcb{Ivan Shuisky} suffer from the same fate as \tcb{Benny Frey}?\\
 \hline
\multicolumn{2}{c}{}\\

\end{tabular}

\begin{tabular}{|p{4cm}|p{11cm}|}
\hline
\multicolumn{2}{|c|}{\textbf{Claim Verification}} \\ 
\hline

 \textbf{Exemplar Original Claim} & The \tcb{Bugs Bunny} cartoons were influenced by the cartoon \tcb{Rick and Morty}.\\ 
 \hline
 \textbf{Exemplar Modified Claim}& \tcb{Giovanni Battista Casti}'s works may be influenced by \tcb{Maria Grazia Lenisa}'s poems. \\
 \hline
\multicolumn{2}{c}{}\\

\end{tabular}

\begin{tabular}{|p{4cm}|p{11cm}|}
\hline
\multicolumn{2}{|c|}{\textbf{Preference Matching}} \\ 
\hline

 \textbf{Exemplar Query} & Sam: I like eating \tcb{pulled} \textcolor{purple}{meats, but not beef or chicken}.\\ 
 \hline
 \textbf{Original Options}& \faCheckSquareO \  \tcb{Shredded} barbecued \textcolor{purple}{pork} shoulder \\
  & \faSquareO \ \textcolor{purple}{Pork} chops made with orange juice, garlic, and thyme\\
& \faSquareO \  Shredded barbecued beef with Worcestershire sauce\\
& \faSquareO \  Sandwiches made with shredded barbecued chicken thighs \\
& \faSquareO \  Chicken, mushroom, and tomato baked in a sauce mixture\\
  
 \hline
 \textbf{Added Option} & \faSquareO \  \tcb{Pulled} \textcolor{purple}{Pork} in a Crockpot with \tcr{garlic} and orange juice\\

 \hline
 \textbf{Personal KG} & (Sam, occupation, painter), (Sam, age, 29), (Sam, medical condition, \tcr{allium allergy}), \ldots\\ &, (Sam, religion, Christianity), (Sam, medical condition, lactose intolerance).\\
 \hline
\end{tabular}
\caption{Exemplar queries form Datasets used for each task and modifications applied to them. Modified queries in Question answering and Claim verification target obscure entities to evaluate robustness to popularity shift. The synthetic KG and the new option add personalization aspect to the Preference Matching task.}
\label{dataset}
\end{table*}

\subsection{Experimental Setup}
We compare $R^3$ against LLM baselines with Chain-of-Thought (CoT) prompting, both in zero-shot~\cite{0-cot} and few-shot ($k=2$) settings ~\cite{cot} to evaluate the need for a KG to answer these queries, and three recent LLM-based KGQA models, KB-BINDER~\cite{kbbinder}, KGR~\cite{kgr}, and KAPING~\cite{kaping}. For question answering and claim verification tasks, we evaluate all methods on both original queries (targeting famous entities) and modified queries (targeting long-tail entities) to study their robustness to popularity shift.  We use GPT-3.5 Turbo as the LLM for all models. In addition to accuracy, we perform human evaluation to measure factual and reasoning faithfulness. In particular, we use FActScore \cite{factscore}, which measures the percentage of atomic facts in an LLM's response supported by a knowledge base, and Reasoning score, which measures the proportion of LLM responses in which there are no logical reasoning errors. For preference matching, our human evaluation consists of measuring \textit{Accuracy of Reasons} which is the fraction of correct answers that were obtained from correct reasons.

\subsection{Results}
\subsubsection{Question Answering}
The results for the question answering task are presented in table \ref{qa-cv-results}. $R^3$ beats all baselines, achieving an accuracy of 0.82 and 0.73 in the original and long-tail settings respectively. Although the strongest baseline, KAPING, achieves comparable accuracy, human evaluation reveals that KAPING’s answers are far less reliable than those of $R^3$. KB-BINDER’s performance is much lower than other methods, because KB-BINDER is a semantic parsing method that only supports factoid queries and not ones that require commonsense reasoning. Although 0-shot and few-shot CoT achieve 0.70 accuracy on the original queries, their accuracies drop significantly in the long-tail setting to 0.32 and 0.43 respectively. We also observe in the long-tail setting a sharp increase in the number of questions for which the LLM responds “I don’t know.”

Among all methods, $R^3$ hallucinates the least, with the highest FActScores, 0.97 and 0.96, in the original and long-tail settings respectively. KAPING’s FActScores, 0.74 and 0.59, are significantly lower than $R^3$, despite leveraging dense retrieval. This is because KAPING’s retrieval is limited to entities in the question, which works only for single-hop queries. 
For multihop queries, KAPING resorts to the LLM’s internal knowledge. From our LLM baselines, we observe low FActScores, indicating that LLM’s internal knowledge is insufficient. In contrast, $R^3$ enforces strict grounding on the KG for reasoning, and has an iterative mechanism for identifying what additional facts are required, which leads to near perfect FActScores. 

Not only are the FActScores of baseline methods significantly lower than $R^3$, but we also observe for all baselines a significant decrease in FActScore on long-tail queries. For instance, KAPING’s FActScore drops by 0.15 from 0.74 to 0.59. These results show that baseline method hallucinations are exacerbated in the long-tail setting due to LLMs being unable to faithfully recall long-tail knowledge. For KAPING, we also observe that the entity linker fails more often to identify long-tail entities, which inevitably leads to ungrounded hallucinated answers in the absence of relevant triples. In contrast, $R^3$ maintains a high FActScore in both the original and long-tail settings with respective scores of 0.97 and 0.96, which indicate its robustness to shifts in entity popularity. 

$R^3$ also maintains the highest reasoning score compared to all baselines, achieving a score of 0.97 and 0.95 in the original and long-tail settings respectively, compared to the next best method, few-shot CoT, which achieves reasoning scores of 0.92 and 0.90. Because $R^3$ makes the commonsense inference process explicit by axiomatically surfacing the commonsense inference rules, $R^3$ provides both more verifiable and faithful chains of reasoning with less errors. In contrast, KAPING has a low reasoning score. We qualitatively observe that due to the low precision of the facts retrieved by KAPING, the LLM is frequently misled by the irrelevant facts. Elsewhere, KGR has the lowest reasoning score. Without CoT, KGR’s initial response often contains poor reasoning, which then leads to poor retrofitting and thus a low FActScore as well. 
Note that we do not perform human evaluation for KB-BINDER since it is a semantic parsing method that outputs SPARQL queries which are incompatible with FActScore and reasoning scores.  

\subsubsection{Claim Verification}
The results for the claim verification task are presented in Table \ref{qa-cv-results}. Although 2-shot CoT beats our method on the original queries, our method is robust in the long-tail setting, achieving the same accuracy as the original setting whereas 2-shot COT’s accuracy drops significantly by 0.51. 

We observe that again $R^3$ maintains the highest FActScore, 0.98, in both the original and long-tail settings. In contrast, similar to the question answering task, all baseline methods have significantly lower FActScores that also decrease significantly in the long-tail setting. The low and decreasing FActScores in both the question answering and claim verification task crucially demonstrate that LLMs suffer from high rates of hallucination which are exacerbated in long-tail settings. 

$R^3$ also maintains the highest reasoning score among all methods, 0.04 better than the next-best method which is few-shot CoT. Interestingly, with few-shot CoT, we qualitatively observe that the LLM at times erroneously follows the reasoning strategies in the examples. We believe that explicitly surfacing commonsense axioms is crucial for correctly guiding the subsequent reasoning. Again, KAPING’s low precision KG retrieval misleads the LLM, resulting in low reasoning scores, and KGR’s poor reasoning leads to suboptimal initial responses that KGR has difficulty retrofitting. 

A statistical analysis of these results is provided in Appendix \ref{sec:appendix-statistical}, which verifies that $R^3$ statistically significantly reduces sources of hallucination on three of the studied datasets. We also provide anecdotal examples of $R^3$'s performance in addressing LLM misbeliefs in Appendix \ref{sec:appendix-override}.

\subsubsection{Preference Matching}
Results of the preference matching task are provided in Table \ref{tab:preference}. Since the personal KG does not support SPARQL queries, KB-BINDER cannot be evaluated on it. KGR and pure LLM baselines also cannot be evaluated on this task since they can only make claims or provide answers about entities that LLMs are aware of, and not about users in a synthetic dataset. So, the only relevant baseline is KAPING. Results of this comparison vividly identify that on the challenging task of personalized preference matching, $R^3$ obtains a considerably higher accuracy. We also observe that the Accuracy of reasons for $R^3$ is more than double the number for KAPING, which again reflects its stronger commonsense reasoning ability due to its special approach for surfacing commonsense axioms.
\begin{table}[]
    \begin{tabular}{|l|c|c|}
    \hline
         \textbf{Method}& \textbf{Accuracy} & \textbf{Accuracy of Reasons}  \\
         \hline
         KAPING & 44 & 31.8\\
         \hline
         $R^3$ & \textbf{57} & \textbf{70.17}\\
         \hline
    \end{tabular}
    \caption{Results of Accuracy and Accuracy of reasons (\%) for preference matching task}
    \label{tab:preference}
\end{table}

\section{Conclusion}
We proposed $R^3$, a novel framework that enables answering KG queries involving commonsense reasoning using LLMs in a verifiable manner by axiomatically surfacing their intrinsic commonsense knowledge. Key experimental results exhibit the efficacy of $R^3$ across different tasks related to KGQA and its superior performance to existing baselines. The promising performance of $R^3$ combined with its verifiability and robustness to entity popularity opens up possibilities for versatile future extension to address broader ranges of tasks and improve the flexibility and accessibility of KGs and reliability of LLM-based reasoners.

\section{Limitations}
While we believe this work has made significant forward progress in leveraging KG content for commonsense question answering (QA), our method $R^3$ (like any QA method) has natural limitations that we hope will encourage further investigation and future work.  

The quality of the reasoning process in $R^3$ relies on the quality of the natural language axioms generated. We observe through our experiments that in cases where the quality of axioms is insufficient, the reasoner is misled resulting in an undetermined answer at the end of the exploration budget identified. 
Furthermore, due to the importance of avoiding hallucination, our model takes a conservative and rigorous approach to ground every factual premise on KG triples. Therefore, our model typically leaves more questions unanswered than other baselines (which we considered 
an incorrect response in calculating the accuracy).

Furthermore, as in most LLM-based models, for having a proper performance, LLM-based components of our model require clear explanation of the task provided in the prompts to them, as well as a number of few-shot examples that clarify the intent of the task description further.

We consider further studies into the above limitations as open areas of future work. Studying the trade-off between rigor and the rate of unanswered questions, as well as studying the robustness of our model to different prompting styles are key research questions that we consider for future.

\section{Ethics Statement}
This work intends to provide a solution for improving the correctness and faithfulness of LLMs in the task of commonsense KGQA. Additionally, it seeks to improve the verifiability of the generated answers, thereby aiding in the detection and mitigation of incorrect or potentially harmful content. However, it is important to acknowledge that this approach (a) relies on LLMs that may hallucinate and (b) presumes the accuracy of the knowledge graph (KG) data and lacks any capacity to correct erroneous or noisy information present within the KG. 
Hence, it is imperative to ensure accuracy of the KG and that the reasoning steps introduced by $R^3$'s LLM are free of both hallucinations and otherwise incorrect, biased, or unethical conclusions that may be harmful to downstream users.

\bibliography{main}
\appendix

\section{Analysis of $R^3$ Components}
\label{sec:appendix-ablation}

\begin{table*}[tbh!]
\centering
\small
\begin{tabular}{@{}lllcccccc@{}}
\toprule
  \multicolumn{3}{l}{}
   &
  \multicolumn{2}{c}{\textbf{Question Answering}} &
  \textbf{} &
  \multicolumn{2}{c}{\textbf{Claim Verification}} \\ \cmidrule(lr){4-5} \cmidrule(l){7-8}
 &
   &
   &
  \textbf{Original} &
  \textbf{Long-Tail} &
   &
  \textbf{Original} &
  \textbf{Long-Tail} \\ \midrule
    &
  Standard Entity linker~\cite{ayoola2022refined} &
   &
  0.938 &
  0.854 &
   &
  0.974 &
  0.986\\ \midrule
 &
  LLM-based Entity extractor &
   &
  0.960 &
  0.979&
   &
  0.928 &
  0.986\\ \midrule
  &
  $R^3$ &
  &
    \textbf{1.00}&
  \textbf{1.00} &
   &
  \textbf{1.00} &
  \textbf{1.00} \\ \bottomrule

\end{tabular}
\caption{Success rate of different approaches in extracting entities from queries of each dataset split. The superior performance of $R^3$ in extracting the relevant entities from queries compared to the ablations shows the importance of both entity extraction modules in the $R^3$ framework.}
\label{tab:entity-linking-methods}
\end{table*}

\begin{table*}[tbh!]
\centering
\small
\begin{tabular}{@{}lllcccccc@{}}
\toprule
  \multicolumn{3}{l}{}
   &
  \multicolumn{2}{c}{\textbf{Question Answering}} &
  \textbf{} &
  \multicolumn{2}{c}{\textbf{Claim Verification}} \\ \cmidrule(lr){4-5} \cmidrule(l){7-8}
 &
   &
   &
  \textbf{Original} &
  \textbf{Long-Tail} &
   &
  \textbf{Original} &
  \textbf{Long-Tail} \\ \midrule
    &
  $R^3$ without Entity Linker &
   &
  0.807 &
  0.713 &
   &
  0.820 &
  0.846 \\ \midrule
 &
  $R^3$ without LLM-based Entity Extractor &
   &
  0.793 &
  0.700 &
   &
  0.827 &
  0.753 \\ \midrule
  &
  $R^3$ &
  &
    \textbf{0.820} &
  \textbf{0.727} &
   &
  \textbf{0.846} &
  \textbf{0.853} \\ \bottomrule

\end{tabular}
\caption{Accuracy of $R^3$ compared to its variants with ablated entity extraction modules. The higher success rate of $R^3$ in extracting queries also results in a higher accuracy.}
\label{tab:entity-linking-accuracies}
\end{table*}

The framework of $R^3$ comprises several integral parts and modules that collectively enhance its performance. In Section \ref{sec:experiments}, we delineated the motivation and function of each component within the $R^3$ framework. To further substantiate the significance of each part and assess its impact on overall performance, we conduct a series of ablation studies and experiments in this section. This analysis contrasts the functionality of each component against alternative design choices, providing deeper insights into the necessity of each element in the $R^3$ architecture.

Utilizing the KG facts and grounding the facts used in reasoning on the KG is a cornerstone of the $R^3$ framework. Ablating the use of KG effectively reduces $R^3$ to the few-shot CoT baselines, which we previously compared in Section \ref{sec:experiments}. There are three major steps in answering a commonsense query based on KG:
\begin{itemize}
    \item Extracting KG entities from the query and obtaining the sub-graph containing the queries.
    \item Identifying the facts that are relevant and useful in answering the question from the extracted sub-graph.
    \item Answering the question using these relevant facts.
    
\end{itemize}

$R^3$ adds a critical step that governs its search process for answering the query, which is surfacing the commonsense axiom. The importance of this step was shown through experiments conducted in Section \ref{sec:experiments}. Removing the surfaced commonsense axioms and the tree-structured search that $R^3$ employs to answer queries simplifies it to KAPING, one of the baseline we evaluated in Section \ref{sec:experiments} and showed that it was outperformed by $R^3$.

In this section, we study the options and design choices that can be considered for each of the three enumerated steps and examine the influence of ablating components utilized in the $R^3$ framework in each step.

\subsection{Obtaining Relevant Sub-graph}
The first step in answering a query in the $R^3$ framework is extracting the KG entities that are targeted in the question to obtain their relevant sub-graph from the KG and answer the query based on it. We consider three design choices for this step:
\begin{itemize}
    \item Using existing entity-linking methodologies
    \item Using an LLM to extract entities from the query
    \item Using a combined approach by uniting entities obtained by these two methods (used in $R^3$)
\end{itemize}

Existing KGQA methodologies often rely on entity-linking techniques~\cite{ li2020efficient, ayoola2022refined} that efficiently extract well-known entities. However, since these methods were not trained on sufficient data from long-tail and recent entities that $R^3$ aims to address, they might not be able to perform successful entity extraction for those queries. To address this possible issue, $R^3$ also leverages an LLM-based entity extractor. In this analysis, we study the role and importance of each of these entity extraction techniques. 

To this end, we first compare the entity extraction performance of each of these entity linking methodologies by using each of them to extract entities for all queries of all subsets of the dataset, and comparing the sets of retrieved entities against the set of ground truth entities that are contained in all queries. Results of this experiment are shown in Table \ref{tab:entity-linking-methods}. In the first row of this table, we use ReFinED, a standard entity linking methodology that is specialized for Wikipedia and Wikidata entities, and in the second row, we just use our LLM-based entity extractor. The final row refers to the final set of entities that we use in $R^3$ which is basically the union of the entities retrieved by each of these methods. From this table, we can verify that although both entity extraction methods have a high success rate in extracting the entities, they are both imperfect and fail to extract a fraction of the entities from some queries. However, when their union is used in $R^3$, all entities can be successfully retrieved to extract their relevant sub-graph. This means that on every query that one of these methods fails to extract the correct entity, the other method successfully compensates for it. We note that this perfect entity extraction result that is obtained for $R^3$ is confined to the datasets that we studied in this paper and across other datasets, there might be cases in which both entity extraction methods fail. However, using both methods considerably increases the chance of successful retrieval. This table also validates our hypothesis that the standard entity linking mythologies may be challenged more in extracting the long-tail entities, but the LLM-based entity extractor is more robust to entities' popularity. 

To further verify the importance of utilizing both sub-graph extraction methodologies, we examine the role of each method in the overall performance of $R^3$. We repeat all experiments for both tasks—question answering and claim verification—while ablating the two entity extraction methodologies. The results of this experiment are presented in Table \ref{tab:entity-linking-accuracies}. These findings underscore the significance of the entity extraction scheme employed in $R^3$. In every case, the combined use of both entity extraction methodologies (as implemented in $R^3$) enhances the accuracy across all tasks. Additionally, this table highlights the contribution of the LLM-based entity extractor introduced in this work to the method's overall performance.

In conclusion, for extracting the relevant sub-graph—a crucial first step in answering commonsense queries based on the factual knowledge of the KG—the combined methodology introduced in $R^3$ outperforms both the classical specialized entity linkers and the standalone use of the LLM-based entity extractor. This conclusion is supported by observations of both the success rate in entity retrieval and the overall query-answering performance.

\begin{table*}[tbh!]
\centering
\small
\begin{tabular}{@{}lllcccccc@{}}
\toprule
  \multicolumn{3}{l}{}
   &
  \multicolumn{2}{c}{\textbf{Question Answering}} &
  \textbf{} &
  \multicolumn{2}{c}{\textbf{Claim Verification}} \\ \cmidrule(lr){4-5} \cmidrule(l){7-8}
 &
   &
   &
  \textbf{Original} &
  \textbf{Long-Tail} &
   &
  \textbf{Original} &
  \textbf{Long-Tail} \\ \midrule
    &
  $R^3$ with truncation instead of pruning&
   &
  0.527 &
  0.480 &
   &
  0.726 &
  0.800 \\ \midrule
 &
  $R^3$&
   &
  \textbf{0.820} &
  \textbf{0.727} &
   &
  \textbf{0.846} &
  \textbf{0.853} \\ \bottomrule

\end{tabular}
\caption{Accuracy of $R^3$ compared to its variant in which semantic similarity-based sub-graph pruning is replaced with truncation. The significant drop in the performance of $R^3$ after ablating the sub-graph pruning approach is due to the loss of essential KG facts due to naive truncation.}
\label{tab:pruning}
\end{table*}

\begin{table*}[tbh!]
\small
\centering
\begin{tabular}{@{}ccccc@{}}
\toprule
  \multicolumn{1}{l}{\textbf{Reasoning Tree Depth}}
   &
  \multicolumn{1}{c}{\textbf{Question Answering}} &
  \textbf{} &
  \multicolumn{1}{c}{\textbf{Claim Verification}} \\ 
\midrule
    0
   &
  0.473 & &
  0.620 \\ \midrule
  1
  &
  0.553 & &
  0.707 \\ \midrule
  2
  &
  0.727 & &
  0.853 \\ \midrule
  3
  &
  0.733 & &
  0.860 \\ \bottomrule

\end{tabular}
\caption{Accuracy of $R^3$ in question answering and claim verification tasks against the depth of reasoning tree generally increases with the increased tree depth. The significant gap between the reasoning depth of $0$ and the reasoning depth of $2$ which is the original $R^3$ results indicates the importance of the iterative mechanism of $R^3$ for answering multi-hop queries.}
\label{tab:depth}
\end{table*}

\subsection{Sub-graph Pruning}
Due to the potentially large size of the relevant sub-graph that is retrieved, it is crucial to prune it to enable the use of an LLM-based reasoner that has a limited context length. However, it is crucial not to prune out the essential KG facts from the relevant sub-graph that are essential in answering the query. We consider two possible approaches in this regard:
\begin{itemize}
    \item Truncating the retrieved sub-graph to fit in the context length.
    \item Using more intelligent approaches such as semantic similarity to identify the more relevant facts.
\end{itemize}

In $R^3$, we used an approach based on the semantic similarity between the commonsense axiom and facts in the relevant sub-graph. In order to verify the efficacy of this approach in preserving the essential KG facts while pruning the irrelevant ones, we perform an experiment in which we ablate this semantic similarity-based approach of sub-graph pruning. However, due to the large size of the retrieved sub-graph, we truncate the set of facts to fit the context size of the LLM. 

The results of comparing the outcome of this sub-graph pruning method against the semantic similarity-based approach used in $R^3$ are presented in Table \ref{tab:pruning}. Evidently, truncating the sub-graph leads to a significant drop in accuracy across all dataset splits, as it often prunes essential facts. These results confirm the necessity of the sub-graph pruning approach employed in $R^3$ for judiciously selecting the facts that are useful in answering the queries.

\subsection{Iterative Process for Answering Multi-hop Queries}
$R^3$ is equipped with a tree-structured search mechanism for answering queries. As illustrated in the workflow of $R^3$ in Figure \ref{fig:diagram}, each branch of the tree undergoes an iterative process of sub-graph retrieval and pruning, attempting to answer the query, and identifying missing information at deeper levels of the tree. This iterative process enables $R^3$ to perform multi-hop reasoning on the KG, thereby providing fact-based answers.

In this experiment, we validate the necessity of the tree-structured search process in answering commonsense queries for question answering and claim verification tasks. To achieve this, we vary the maximum depth of the search tree and conduct experiments on the long-tail subsets of the question answering and claim verification datasets. Results of this experiment are presented in Table \ref{tab:depth}.

We first completely ablate this iterative process and try to answer queries on the first try. Results of this experiment are shown in the first row ($\text{depth} = 0$) which shows a considerably lower accuracy than the original $R^3$ performance that we reported in the paper using $\text{depth} = 2$. By increasing the tree depth which is equivalent to an increased number of iterations for performing multi-hop reasoning, the accuracy in both tasks increases, until it plateaus at $\text{depth} = 2$ as there are limited queries requiring more reasoning steps on these two datasets.

The results of this study underscore the critical importance of the iterative process for effectively answering multi-hop commonsense queries within the $R^3$ framework.

\section{Statistical Analysis}
\label{sec:appendix-statistical}

 \begin{table*}[tbh!]
\centering
\small
\begin{tabular}{@{}lllcccccccc@{}}
\toprule
  \multicolumn{3}{l}{}
   &
  \multicolumn{2}{c}{\textbf{Accuracy}} &
  \textbf{} &
  \multicolumn{2}{c}{\textbf{FActScore}} &
  \multicolumn{1}{l}{} &
  \multicolumn{2}{c}{\textbf{Reasoning}} \\ \cmidrule(lr){4-5} \cmidrule(lr){7-8} \cmidrule(l){10-11} 
 &
   &
   &
  \textbf{Original} &
  \textbf{Long-Tail} &
   &
  \textbf{Original} &
  \textbf{Long-Tail} &
  \multicolumn{1}{l}{} &
  \textbf{Original} &
  \textbf{Long-Tail} \\ \midrule
    &
  p-value &
   &
  0.1868 &
  0.1443 &
   &
  0.0004 &
  0.00007 &
  \multicolumn{1}{l}{} &
  0.0290 &
  0.0606 \\ \midrule
 &
  Best Baseline &
   &
  KAPING &
  KAPING &
   &
  KAPING &
  KAPING &
  \multicolumn{1}{l}{} &
  2-shot CoT &
  2-shot CoT

 \\ \bottomrule
\end{tabular}
\caption{Results of the statistical tests between the outputs of $R^3$ and the best-performing baseline across all queries per column. }
\label{tab:statistical}
\end{table*}

In order to evaluate the statistical significance of the superior performance of $R^3$ in comparison to the baselines that were reported in Table \ref{qa-cv-results} of Section \ref{sec:experiments}, we conducted a statistical test. Each subset of this dataset contains 150 queries, resulting in a total of 600 queries across the two tasks with the original and long-tail settings.

In this test, we consider queries of each column with the responses provided by $R^3$ and answers given by the best-performing baseline across all queries per column, resulting in a total of 300 query-answer pairs for each column. Since FActScore is a numerical metric, we employed the paired t-test to obtain the statistical significance, while for the Accuracy and Reasoning metrics, we utilized McNemar's test (also for paired data) considering the binary nature of the data. We also tried calculating the Fisher's exact test and it provided much more favorable p-values indicating a stronger significance of $R^3$'s superiority, but we do not believe it is appropriate for this paired comparison of each method on the same queries and therefore, do not include its results.

Results of the p-values reflecting the statistical significance test are presented in Table \ref{tab:statistical}. While the p-values are not high enough to make strong statistical claims that $R^3$ performs statistically significantly better than the best baseline in terms of Accuracy, we note that the purpose of “Right for the Right Reasons” ($R^3$) is to maintain the accuracy of existing state-of-the-art QA methods while reducing fact and reasoning hallucinations. Fact and reasoning hallucinations are respectively measured by the FActScore and Reasoning metrics. On these metrics, p-value results show very encouraging statistical results. Reasoning results for $R^3$ appear significantly better for the Original versions of the datasets (p-value < 0.05) and just miss the 0.05 significance level for the Long-tail version by a small margin. Critically, for the FActScore, $R^3$ outperforms the best baseline with high statistical significance (p-value < 0.001) for both the Original and Long-tail variants of our datasets indicating a highly statistically significant reduction in fact hallucination for $R^3$.

In summary, this statistical analysis shows that $R^3$ potentially outperforms and at least matches the Accuracy of state-of-the-art methods and it statistically significantly reduces sources of hallucination on three out of the four datasets (and almost significantly on the fourth).

\section{Overriding LLM Misbeliefs with KG Facts}
\label{sec:appendix-override}
During their training process, LLMs acquire substantial factual knowledge about various objects and entities. However, as observed in the experimental results presented in Section \ref{sec:experiments}, utilizing their internal knowledge in answering commonsense queries is prone to hallucination, especially in answering queries about long-tail entities. $R^3$ addresses this challenge by grounding its answers on the facts from the KG that are more reliable. In fact, it is likely that the internal knowledge of the LLM disagrees with a fact that $R^3$ obtains from the KG, and in these circumstances, $R^3$ overrides the LLM’s potentially mistaken belief with a credible KG fact, which resulted in higher accuracy and factual correctness of the $R^3$’s responses compared to the LLM-based baselines.

In order to verify that  $R^3$ succeeds in overriding mistaken beliefs of the LLM by correct and relevant facts from the KG and observe the contribution of this approach to the superior performance of $R^3$, in this section, we provide anecdotal examples of responses provided by $R^3$ and the LLM-based baselines to 30 queries. These queries are chosen from the subsets of the benchmark datasets on which $R^3$ outperforms the baselines, i.e., question answering task in both original and long-tail settings and the claim verification task in the long-tail setting. We choose these queries among the queries that the LLM bases its answers on facts and does not respond by just \textit{``Yes’’}, \textit{``No’’}, or \textit{``I don’t know’’}. We also provide the complete sets of responses provided by $R^3$ and all baselines on all queries in our repository.

Each anecdotal example is provided in a table that contains the query, the correct answer, and the set of responses provided by each method followed by a brief discussion comparing those responses. Correct facts that are used in each response are indicated by green text colour and incorrect ones are shown in red. Also, incorrect reasoning steps are indicated by red underline.

\onecolumn
\clearpage
\paragraph{Question Answering: Long-tail}

    






\begin{longtable}{| p{0.95\textwidth} |}
\hline
\textbf{Query:} Could someone in Vaughan take a taxi to the Jamjuree Art Gallery? (No)\\
\hline
\textbf{0-Shot CoT:} First, let's find out if there are taxis available in Vaughan. In Vaughan, taxis are available for local transportation. Next, let's check the location of the Jamjuree Art Gallery to see if it's within a reasonable distance from Vaughan. The Jamjuree Art Gallery is located in \tcr{Toronto}, which is adjacent to Vaughan. With this information, it is reasonable to conclude that someone in Vaughan could take a taxi to the Jamjuree Art Gallery, given the proximity of the locations. Final answer: Yes.

\\

\textbf{2-Shot CoT:} The Jamjuree Art Gallery is located in \tcg{Bangkok, Thailand}. Vaughan is a city in \tcg{Ontario, Canada}. Since they are in different countries and continents, it is not feasible to take a taxi from Vaughan to the Jamjuree Art Gallery. Final Answer: No.
\\
\\
\textbf{$\mathbf{R^3}$:} Based on the general rule, the Jamjuree Art Gallery must be within a reasonable distance from Vaughan and accessible by road for someone in Vaughan to take a taxi to it. Fact “(Jamjuree Art Gallery, country, Thailand)'' tells us that the Jamjuree Art Gallery is located in \tcg{Thailand}, which is quite far from \tcg{Vaughan, Ontario, Canada}. Fact “(Vaughan, is, city in Ontario, Canada)'' tells us that Vaughan is a city in Ontario, Canada. Although the facts don't provide explicit information about the distance between Vaughan and the Jamjuree Art Gallery, we can make a plausible assumption based on general knowledge. Thailand is located on the other side of the world from Canada, and it is not within a reasonable distance for someone in Vaughan to take a taxi to the Jamjuree Art Gallery. Therefore, the answer is: No, someone in Vaughan cannot take a taxi to the Jamjuree Art Gallery.
\\
\\
\hline
In the 0-shot response, the LLM has hallucinated about the location of the Jamjuree Art Gallery and believed that it is located in Toronto. Although the reasoning about the possibility of traveling by taxi between two adjacent cities is correct, the hallucinated fact has led to an incorrect final answer. The 2-shot response is correct and does not include this hallucination. The response provided by $R^3$ is completely correct and based on KG facts. Also, it points to the specific KG facts used in its reasoning which allows for verification of the response.\\

\hline

\end{longtable}
\begin{longtable}{| p{0.95\textwidth} |}
\hline
\textbf{Query:} Is the ``It's Hyorish'' an older album than ``Nasime vasl''? (No)\\
\hline
\textbf{0-Shot CoT:} First, let's determine the release dates of the albums ``It's Hyorish'' and ``Nasime vasl''. Once we have this information, we can compare the release dates to see which album is older. The album ``It's Hyorish'' \tcr{by Nasimevasl} was released on \tcr{October 30, 2019}. On the other hand, ``Nasime vasl'' \tcr{by Marjan Farsad} was released on \tcr{November 23, 2017}. Comparing the release dates, ``Nasime vasl'' is older than ``It's Hyorish''. Final answer: Yes, ``Nasime vasl'' is an older album than ``It's Hyorish''.\\
\\
\textbf{2-Shot CoT:} The album ``It's Hyorish'' was released \tcr{in 1990}, while ``Nasime vasl'' was released \tcr{in 1978}. Therefore, ``It's Hyorish'' is not older than ``Nasime vasl''. Final Answer: No.\\
\\
\\
\textbf{$\mathbf{R^3}$:} Based on the general rule, we need to compare the publication dates of ``It's Hyorish'' and ``Nasime Vasl'' to determine which album is older. Fact “(It's Hyorish, publication date, 2008)” tells us that "It's Hyorish" was published \tcg{in 2008}, and Fact ``(Nasime Vasl, publication date, 2003/05/21)'' tells us that ``Nasime Vasl'' was also published \tcg{in 2003}. Comparing the publication dates, we can see that ``Nasime Vasl'' was published earlier than ``It's Hyorish''. Therefore, the answer is: No, ``Nasime Vasl'' is an older album than ``It's Hyorish''.\\
\\
\hline
In the 0-shot CoT response, the LLM makes multiple factual hallucinations. First, it states that ``It's Hyorish'' is an album by ``Nasimevasl'', whereas such singer does not exist and ``Nasime vasl'' is the name of another album. It also states that ``It's Hyorish'' was released on October 30, 2019 which is incorrect. It also states that ``Nasime vasl'' is an album by ``Marjan Farsad'' and that ``Nasime vasl'' was released on November 23, 2017, which are both incorrect. Also, in the 2-shot CoT response, the LLM hallucinated about the dates the two albums were released. However, $R^3$ correctly overrides the mistaken beliefs of the LLM by KG facts and provides the correct answer.
\\
\hline
\end{longtable}

\begin{longtable}{| p{0.95\textwidth} |}
\hline
\textbf{Query:} Did Ilary Blasi celebrate a silver anniversary with Francesco Totti? (No)\\
\hline
\textbf{0-Shot CoT:} To the best of my knowledge, Ilary Blasi \tcr{did celebrate} a silver anniversary with Francesco Totti. ``Silver anniversary'' usually refers to the 25th anniversary, and in this context could imply a significant milestone in their relationship. Ilary Blasi and Francesco Totti are \tcg{known public figures} and have been married for many years, so it is possible that they celebrated a silver anniversary.\\
\\

\textbf{2-Shot CoT:} Yes. Ilary Blasi and Francesco Totti \tcr{celebrated their silver (25th) anniversary in 2020}. Final answer: Yes.
\\
\\

\textbf{$\mathbf{R^3}$:} According to the general rule provided, Ilary Blasi and Francesco Totti must have been married for 25 years to celebrate a silver anniversary together. Fact “(Francesco Totti, spouse, Ilary Blasi; start time, 2005; end cause, divorce; end time, 2022)” states that Ilary Blasi and Francesco Totti were married \tcg{from 2005 to 2022}, which is a total of 17 years, not 25 years. Therefore, based on the information given, Ilary Blasi and Francesco Totti did not celebrate a silver anniversary together. Hence, the answer is: No, Ilary Blasi did not celebrate a silver anniversary with Francesco Totti.\\
\\
\hline
Both 0-Shot and 2-Shot responses hallucinate about the length of Ilary Blasi and Francesco Totti's marriage. However, ${R^3}$ overrides the mistaken belief by grounding its response on the KG fact about their divorce which leads to correctly calculating their length of marriage.\\
\hline

\end{longtable}

\begin{longtable}{| p{0.95\textwidth} |}
\hline
\textbf{Query:} Is it likely for the director of Demitri Cantemir to fast during Ramadan? (No) \\
\hline
\textbf{0-Shot CoT:} First, \ul{Demitri Cantemir} was a \tcg{Moldavian historian, composer, and scholar of Ottoman history and music}. It is known that he was of \tcg{Orthodox Christian} faith, which doesn't require fasting during Ramadan. So, it is unlikely for the director of Demitri Cantemir to fast during Ramadan. Final answer: No.\\
\\
\textbf{2-Shot CoT:} \ul{Demitri Cantemir} was a \tcg{Moldavian philosopher, historian, composer, and a man of arts}. He was also a \tcr{convert to Islam} and \tcg{lived in Constantinople} for a portion of his life. Since Ramadan is an important month of fasting for Muslims, it is likely that Demitri Cantemir, being a Muslim, observed fasting during Ramadan. Final answer: Yes.
\\
\\
\textbf{$\mathbf{R^3}$:} The general rule states that the likelihood of the director of a film fasting during Ramadan depends on whether the director observes the Islamic faith. Fact “(Demitri Cantemir, director, Vlad Ioviţă)” tells us that \tcg{Vlad Ioviţă} is the director of Demitri Cantemir. Fact “(Vlad Ioviţă, religion or worldview, Eastern Orthodox Church)” tells us that Vlad Ioviţă's religion is \tcg{Eastern Orthodox Church}, not Islam. Since Vlad Ioviţă does not follow the Islamic faith, it is not likely for him to fast during Ramadan. Therefore, the answer is: No, it is not likely for the director of Demitri Cantemir to fast during Ramadan.\\
\\
\hline
The 0-Shot answer includes a reasoning error. Although the facts it provides about who Demitri Cantemir was and the religion that he believed in are correct, the fact that Demitri Cantemir followed Christianity does not imply that the director of a movie about him is not likely to fast. Also, the response provided by 2-Shot CoT contains the same reasoning error by using information about Demitri Cantemir's religous beliefs to judge about the religous beliefs of the director of the movie about him. This response also includes a hallucinated fact about Demitri Cantemir's conversion to Islam. The response provided by $R^3$ fixes both errors. First, it has a mechanism for identifying the required missing information that allows it to answer multi-hop questions such as this one. $R^3$ determines that information about Demitri Cantemir's director is required and correctly retrieves the required information from the KG and uses it to generate the response. Second, the response is grounded on the KG facts and does not contain hallucinations as the one in the 2-shot CoT. \\

\hline
\end{longtable}

\begin{longtable}{| p{0.95\textwidth} |}
\hline
\textbf{Query:} Was Darya Dadvar from the same city that used to be Afsharid Empire’s center of power? (Yes)\\
\hline
\textbf{0-Shot CoT:} The capital of the Afsharid Empire was \tcg{Mashhad}. Darya Dadvar was born in \tcg{Mashhad}. Therefore, the answer is Yes.\\
\\
\textbf{2-Shot CoT:} Darya Dadvar was born in \tcr{Rasht, Iran}. The center of power of the Afsharid Empire was \tcg{in Mashhad}. Since Rasht and Mashhad are different cities, Darya Dadvar is not from the same city that used to be the center of power of the Afsharid Empire. Final Answer: No.
\\
\\
\textbf{$\mathbf{R^3}$:} Based on General Rule “Darya Dadvar must have been born in the city that used to be the center of power of the Afsharid Empire to be from the same city.”, Darya Dadvar must have been born in the city that used to be the center of power of the Afsharid Empire to be from the same city. Fact ``(Darya Dadvar, place of birth, Mashhad)'' tells us that Darya Dadvar was born \tcg{in Mashhad}. Fact ``(Afsharid Empire, capital, Mashhad)'' tells us that \tcg{Mashhad} was the capital of the Afsharid Empire. Therefore, the answer is: Yes, Darya Dadvar is from the same city that used to be the center of power of the Afsharid Empire, which is Mashhad.\\
\\
\hline
The response provided by 0-Shot CoT is correct, but the one provided by 2-Shot CoT, contains a factual hallucination about the birthplace of Darya Dadvar which shows that LLM is unreliable in answering. However, $R^3$ provides a response which is grounded on the KG and overrides the incorrect belief that was reflected in 2-Shot CoT's response.\\
\hline

\end{longtable}

\begin{longtable}{| p{0.95\textwidth} |}
\hline
\textbf{Query:} Does Toyoharu Tsutsui's alma mater have more history than Newfoundland and Labrador? (Yes)\\
\hline
\textbf{0-Shot CoT:} 1. Toyoharu Tsutsui's alma mater is \tcr{Seijo University}, which was founded \tcr{in 1943}. 2. Newfoundland and Labrador is a \tcg{province in Canada}, known for its \ul{rich history} dating back thousands of years, including Indigenous peoples, Viking settlements, and European exploration. Based on this information, it is clear that Newfoundland and Labrador has significantly more history than Seijo University.\\
\\
\textbf{2-Shot CoT:} Toyoharu Tsutsui's alma mater is \tcr{Tokyo University}, which was established \tcg{in 1877}. Newfoundland and Labrador has a long history dating back \tcr{thousands of years}, including Indigenous occupation, Viking settlements, and European exploration. Therefore, Newfoundland and Labrador has a longer history than Tokyo University. Final answer: No.
\\
\\

\textbf{$\mathbf{R^3}$:} To determine whether Toyoharu Tsutsui's alma mater has more history than Newfoundland and Labrador, we need to compare the histories of both. Fact ``(Newfoundland and Labrador, inception, 1949/03/31)'' tells us that Newfoundland and Labrador was \tcg{established in 1949}. On the other hand, Fact ``(Harvard University, inception, 1636/09/08)'' tells us that \tcg{Harvard University} was \tcg{founded in 1636}. Therefore, based on the information provided, Harvard University has a longer history than Newfoundland and Labrador. Therefore, the answer is: Yes, Harvard University has more history than Newfoundland and Labrador.\\
\\
\hline
In the response provided by 0-Shot CoT, the LLM hallucinates about Toyoharu Tsutsui's alma mater, and also the date Seijo University was founded. Also, there are hallucinations about Toyoharu Tsutsui's alma mater, and the date Tokyo University was founded in 2-Shot CoT. In the response provided by ${R^3}$, these hallucinations are correctly overridden by the respective KG facts, and the final provided answer is also corrected.\\
\hline

\end{longtable}

\begin{longtable}{| p{0.95\textwidth} |}
\hline
\textbf{Query:} Could people have read Mjegull during the Perak war? (No)\\
\hline
\textbf{0-Shot CoT:} To answer the question, we can take a step by step approach. First, let's determine the time period of the Perak war and the publication date of Mjegull. Then we can investigate if there were any circumstances that could have allowed people to read Mjegull during the Perak war. The Perak War occurred from \tcg{1875 to 1876}. Meanwhile, Mjegull was published \tcr{in 1884}. Based on the timeline, it is unlikely that people could have read Mjegull during the Perak war, as the publication of Mjegull came after the war. Final answer: No.\\
\\
\textbf{2-Shot CoT:} Mjegull is a \tcr{fictional language used in the video game "The Witcher 3: Wild Hunt"}. As the Perak War took place in the \tcg{late 19th century}, it is impossible for people to have read Mjegull during that time because it is a creation for a modern video game. Final answer: No.
\\
\\
\textbf{$\mathbf{R^3}$:} The general rule states that people during the Perak War must have had access to the novel Mjegull, which was published in 2009, in order to have read it during that time. Fact “(Perak War, end time, 1876)” tells us that the Perak War ended \tcg{in 1876}, while Fact ``(Mjegull, publication date, 2009; place of publication, Tirana)'' states that the novel Mjegull was published \tcg{in 2009}. Since the publication date of Mjegull is much later than the Perak War, it is not possible for people during the Perak War to have read Mjegull. Therefore, the answer is: No, people could not have read Mjegull during the Perak War.\\
\\
\hline
The 0-Shot CoT response includes a factual hallucination about the time Mjegull was published. Also, 2-Shot CoT response includes an error about the Mjegull entity and attributes it to a fictional language whereas Mjegull here refers to a book. By grounding the response on the KG, $R^3$ succeeds in providing a correct answer to the question which is correct in terms of reasoning and factuality.\\
\hline
\end{longtable}

\begin{longtable}{| p{0.95\textwidth} |}
\hline
\textbf{Query:} Did Ivan Shuisky suffer the same fate as Benny Frey? (Yes)\\
\hline
\textbf{0-Shot CoT:} First, let's identify the fate of Benny Frey. Benny Frey was \tcr{a character in the television show "Hannibal" who met a tragic end}. Now, let's determine the fate of Ivan Shuisky. Ivan Shuisky was a \tcg{historical figure in the 16th century} who was a key figure in Russian politics and was eventually \tcr{executed}. Therefore, the fates of Ivan Shuisky and Benny Frey are different. Final answer: No.\\
\\
\textbf{2-Shot CoT:} Ivan Shuisky and Benny Frey \tcr{are not historical figures} nor are they commonly known individuals. Therefore, I don't have information to determine if they suffered the same fate. Final answer: I don't know.
\\
\\
\textbf{$\mathbf{R^3}$:} According to the general rule, Ivan Shuisky and Benny Frey must have the same cause of death for them to have suffered the same fate. Fact “(Ivan Shuisky, cause of death, carbon monoxide poisoning)” states that Ivan Shuisky's cause of death was \tcg{carbon monoxide poisoning}, and Fact “(Benny Frey, cause of death, carbon monoxide poisoning)” states that Benny Frey's cause of death was also \tcg{carbon monoxide poisoning}. Therefore, since both Ivan Shuisky and Benny Frey had the same cause of death, they suffered the same fate. Therefore, the answer is: Yes, Ivan Shuisky suffered the same fate as Benny Frey.\\
\\
\hline
The 0-Shot CoT response includes an error in identifying who Benny Frey was. Although it correctly identifies who Ivan Shuisky was, it hallucinates about Ivan Shuisky's manner of death. In 2-Shot CoT response, the LLM refrains from answering by stating that it is not familiar with the two entities and their fates. The response provided by $R^3$ utilizes the KG to identify the two entities and obtain information about their manner of death, therefore leading to a correct and grounded response.\\
\hline
\end{longtable}

\begin{longtable}{| p{0.95\textwidth} |}
\hline
\textbf{Query:} Could Kaká have bought a car from Jordan Motor Car Company? (No) \\
\hline
\textbf{0-Shot CoT:} To determine if Kaká could have bought a car from Jordan Motor Car Company, we need to consider the location and timeline. Kaká, the \tcg{Brazilian footballer, played for Orlando City SC in the MLS from 2014 to 2017}, and Jordan Motor Car Company is based \tcr{in Jordan}. Therefore, it is possible that he could have purchased a car from this company if he had been in Jordan during that time. Final Answer: Yes.\\
\\
\textbf{2-Shot CoT:} Kaká, the \tcg{Brazilian footballer}, could have bought a car from Jordan Motor Car Company if the company produced and sold cars while he was active and interested in purchasing one. However, as far as I am aware, there is no public information suggesting Kaká purchased a car from Jordan Motor Car Company. Therefore, without specific information, I don't know if Kaká could have bought a car from Jordan Motor Car Company. Final Answer: I don't know.\\
\\
\textbf{$\mathbf{R^3}$:} Based on General Rule “Jordan Motor Car Company must have been active during Kaká's lifetime for him to have potentially bought a car from them.", Jordan Motor Car Company must have been active during Kaká's lifetime for him to have potentially bought a car from them. Fact ``(Jordan Motor Car Company, inception, 1916)'' tells us that Jordan Motor Car Company was \tcg{founded in 1916} and Fact ``(Jordan Motor Car Company, dissolved, abolished or demolished date, 1931/04/00)'' tells us that it was \tcg{dissolved in 1931}. Fact ``(Kaká, is, Brazilian footballer, born 1982)'' tells us that Kaká was \tcg{born in 1982}, which means he was born after the dissolution of Jordan Motor Car Company. Therefore, Jordan Motor Car Company was not active during Kaká's lifetime. Therefore, the answer is: No, Kaká could not have bought a car from Jordan Motor Car Company.\\
\\
\hline
The response provided by 0-Shot CoT includes a factual hallucination about the location of Jordan Motor Car Company which was in fact located in the United States and not Jordan. The 2-Shot CoT is not able to give a certain answer to the question because it is not aware of the car company's activity period. However, $R^3$ succeeds in providing a correct answer by retrieving the required information from the KG and correctly reasoning over the obtained information.\\
\hline
\end{longtable}

\begin{longtable}{| p{0.95\textwidth} |}
\hline
\textbf{Query:} Could Liau Hiok-hian and Virginia Raggi have met while they both held the position of council member? (No)\\
\hline
\textbf{0-Shot CoT:} First, let's establish who Liau Hiok-hian and Virginia Raggi are. Liau Hiok-hian is a \tcr{Malaysian economist and politician}, known for \tcr{his involvement with the Malaysian Chinese Association (MCA)}. Virginia Raggi is an \tcg{Italian politician who served as the Mayor of Rome}. Given that \ul{Liau Hiok-hian is from Malaysia and Raggi is from Italy}, it is unlikely that they would have met during their time as council members, as they would have been holding positions in different countries. Therefore, it is improbable that they could have met while both holding the position of council member. Final answer: No.\\
\\
\textbf{2-Shot CoT:} Liau Hiok-hian is a \tcr{council member in Malaysia}, while Virginia Raggi is a \tcg{council member in Italy}. \ul{Since they are in different countries}, it is unlikely that they could have met while holding the position of council member. Final answer: No.\\
\\
\textbf{$\mathbf{R^3}$:} Based on General Rule “Liau Hiok-hian and Virginia Raggi must have served as council members during overlapping time periods to have the opportunity to meet while holding the position of council member.” to determine if Liau Hiok-hian and Virginia Raggi could have met while they both held the position of council member, we need to check if their time periods as council members overlap. Fact “(Virginia Raggi, position held, council member; start time, 2013/06/10; end time, 2015/10/30; of, Rome)” states that \tcg{Virginia Raggi held the position of council member in Rome from June 10, 2013, to October 30, 2015}. Fact “(Liau Hiok-hian, position held, Member of Yunlin County Council; start time, 2018/12/25; elected in, 2018 Yunlin County Council election; parliamentary group, New Power Party; electoral district, 1st Constituency of Regional Councilors of Yunlin County; parliamentary term, 19th Yunlin County Council; end time, 2022/12/24)” states that Liau Hiok-hian held the position of \tcg{Member of Yunlin County Council from December 25, 2018, to December 24, 2022}. Since their time periods as council members do not overlap, Liau Hiok-hian and Virginia Raggi could not have met while they both held the position of council member. Therefore, the answer is: No, Liau Hiok-hian and Virginia Raggi could not have met while they both held the position of council member.\\
\\
\hline
Responses of both 0-Shot CoT and 2-Shot CoT contain factual hallucinations about Liau Hiok-hian. They also contain a reasoning error by stating that politicians from different countries are unlikely to meet. However, $R^3$ grounds its reasoning on the KG and overrides the LLM's incorrect beliefs about Liau Hiok-hian by facts from the KG and also provides a correct commonsense reasoning performance by surfacing the commonsense axiom from the LLM.\\
\hline
\end{longtable}

\paragraph{Question Answering: Original}

\begin{longtable}{| p{0.95\textwidth} |}
\hline
\textbf{Query:} Was Dorothea Wendling from same place Porsche originated? (Yes)\\
\hline
\textbf{0-Shot CoT:} To determine if Dorothea Wendling was from the same place as the origin of Porsche, we need to find out where Dorothea Wendling was from and then compare it to the place where Porsche originated. Initial research shows that Dorothea Wendling was born in \tcr{Augsburg, Germany}. Porsche's origins can be traced back to \tcr{the city of Maffersdorf} (now located in the Czech Republic) where Ferdinand Porsche founded his company. Therefore, based on this information, Dorothea Wendling was not from the same place as the origin of Porsche. Final answer: No.\\
\\
\textbf{2-Shot CoT:} Dorothea Wendling, \tcg{a renowned opera singer}, was born in \tcr{Reims, France}, while the Porsche company was founded in \tcg{Stuttgart, Germany}. There is no apparent connection between the birthplaces of Dorothea Wendling and the origin of Porsche. Final answer: No.\\
\\
\textbf{$\mathbf{R^3}$:} The general rule “Dorothea Wendling must have been born in the same place where Porsche originated for them to be from the same place.” states that Dorothea Wendling must have been born in the same place where Porsche originated for them to be from the same place. Fact “(Dorothea Wendling, place of birth, Stuttgart)” tells us that Dorothea Wendling was born in \tcg{Stuttgart}. Fact “(Porsche, location of formation, Stuttgart)” tells us that Porsche was originated in \tcg{Stuttgart} as well. Therefore, based on the information provided, Dorothea Wendling was indeed from the same place where Porsche originated. Therefore, the answer is: Yes.\\
\\
\hline
The 0-Shot CoT response contains factual hallucinations about the birthplace of Dorothea Wendling and the place were Porsche originated from. Likewise, 2-Shot CoT contains a hallucination about the birthplace of Dorothea Wendling. These hallucinated facts are correctly overridden by proper KG facts in the response provided by $R^3$.\\
\hline
\end{longtable}

\begin{longtable}{| p{0.95\textwidth} |}
\hline
\textbf{Query:} Could Carl Friedrich Gauss speak to someone 100 miles away? (No)\\
\hline
\textbf{0-Shot CoT:} To communicate with someone 100 miles away, Carl Friedrich Gauss would have needed to use a method of long-distance communication such as a letter, messenger, or a telegraph, as he lived \tcg{before the invention of modern telecommunications}. Gauss could not have spoken directly to someone 100 miles away without the aid of technology. Final answer: No.\\
\\
\textbf{2-Shot CoT:} Carl Friedrich Gauss, \tcg{a mathematician and scientist}, is known for his work in many areas, \tcg{including mathematics, physics, and astronomy}. While he made important contributions to the understanding of \tcg{electromagnetism} and could theoretically have transmitted messages over long distances with the right equipment, there is no historical record of him actually demonstrating the ability to speak to someone 100 miles away. Final answer: I don't know.
\\
\\
\textbf{$\mathbf{R^3}$:} According to the general rule, Carl Friedrich Gauss must have lived during a time when long-distance communication technology allowed speaking to someone 100 miles away to be able to do so. Fact “(Carl Friedrich Gauss, is, German mathematician and physicist (1777–1855))” tells us that Carl Friedrich Gauss \tcg{lived from 1777 to 1855}. During this time period, long-distance communication technology did not exist in a way that would allow speaking to someone 100 miles away. Therefore, based on the general rule and the time period in which Carl Friedrich Gauss lived, the answer is: No, Carl Friedrich Gauss could not speak to someone 100 miles away.\\
\hline
All three responses are correct. However, the response provided by $R^3$ is more specific by mentioning the exact used fact (date of birth and death of Carl Friedrich Gauss) that makes the response easily verifiable. \\
\hline
\end{longtable}

\clearpage

\begin{longtable}{| p{0.95\textwidth} |}
\hline
\textbf{Query:} Does Adam Sandler skip celebrating Easter? (Yes)
\\
\hline
\textbf{0-Shot CoT:} First, we should consider whether Adam Sandler celebrates Easter at all. There is no public information indicating that he does not celebrate Easter. Additionally, many people of various backgrounds and beliefs choose to celebrate Easter in different ways, so it wouldn't be accurate to assume that he doesn't celebrate it. Therefore, as there is no evidence to suggest that Adam Sandler skips celebrating Easter, the most appropriate answer is: No.\\
\\
\textbf{2-Shot CoT:} As far as I know, there is no information indicating that Adam Sandler skips celebrating Easter. It is possible that he may or may not celebrate Easter, but this information is not readily available. Final Answer: I don't know.\\
\\
\textbf{$\mathbf{R^3}$:} The general rule tells us that to determine if Adam Sandler skips celebrating Easter, we need to consider his religious beliefs and practices. Fact “(Adam Sandler, religion or worldview, Judaism)” states that Adam Sandler is \tcg{Jewish}. Easter is a \tcg{Christian holy day} that commemorates the resurrection of Jesus. Based on the fact that Adam Sandler is Jewish, it is likely that he does not celebrate Easter as it is not a part of the Jewish religious tradition. Therefore, the answer is: Yes, Adam Sandler most likely skips celebrating Easter due to his Jewish faith.\\
\hline
Since LLM is not familiar with Adam Sandler, it cannot provide the correct answer in either the 0-Shot or 2-Shot CoT settings. However, $R^3$ is able to correctly identify the relevant KG facts about his religious beliefs and provide the correct final answer.\\

\hline
\end{longtable}

\begin{longtable}{| p{0.95\textwidth} |}
\hline
\textbf{Query:} Did Melania Trump have same profession as Olga Kurylenko? (Yes)
\\
\hline
\textbf{0-Shot CoT:} Melania Trump was \tcg{a model} before becoming \tcg{the First Lady of the United States}, while Olga Kurylenko is an \tcg{actress} known for her roles in movies. So, \tcr{they did not have the same profession}. Final Answer: No.
\\
\textbf{2-Shot CoT:} No, Melania Trump and Olga Kurylenko \tcr{have different professions}. Melania Trump is a \tcg{former model} and \tcg{the former First Lady of the United States}, while Olga Kurylenko is \tcg{an actress}. Final answer: No.
\\
\textbf{$\mathbf{R^3}$:} According to the general rule, Melania Trump and Olga Kurylenko must share at least one common profession to have the same profession. Facts “(Melania Trump, occupation, jewelry designer)”, “(Melania Trump, occupation, model)”, “(Melania Trump, occupation, celebrity)”, and “(Melania Trump, occupation, businessperson)” tell us that Melania Trump's professions include jewelry designer, model, celebrity, and businessperson. Fact “(Olga Kurylenko, occupation, model)”, Fact “(Olga Kurylenko, occupation, actor)”, Fact “(Olga Kurylenko, occupation, film actor)”, and Fact “(Olga Kurylenko, occupation, voice actor)” tell us that Olga Kurylenko's professions include model, actor, film actor, and voice actor. Since both Melania Trump and Olga Kurylenko \tcg{share the profession of a model}, they have at least one common profession. Therefore, the answer is: Yes, Melania Trump had the same profession as Olga Kurylenko.\\
\hline
The answers provided by both 0-Shot and 2-Shot CoT methods are incorrect because the LLM does not know that Olga Kurylenko is also a model. By obtaining the facts about all of her occupations from the KG, $R^3$ can provide the correct answer.
\\
\hline

\end{longtable}

\begin{longtable}{| p{0.95\textwidth} |}
\hline
\textbf{Query:} Can I hold Bing in a basket? (No)\\
\hline
\textbf{0-Shot CoT:} First, we need to clarify what "Bing" refers to. \ul{Assuming Bing as a small pet or object}, it would depend on the size of Bing and the type of basket. If Bing is small and the basket is appropriately sized and safe for Bing, then it might be possible to hold Bing in a basket. However, if Bing is an object or it's not safe or suitable for Bing to be held in a basket, then it wouldn't be a good idea. Final answer: It depends. If Bing is small and the basket is suitable, then possibly, yes. If not, then no.\\
\\
\textbf{2-Shot CoT:} Bing is a \tcg{search engine owned by Microsoft} and exists as a digital platform. So, no, you cannot hold Bing in a basket. Final answer: No.\\
\\
\textbf{$\mathbf{R^3}$:} The general rule states that Bing in this context refers to Microsoft Bing, which is an online platform and \tcg{not a physical object} that can be held in a basket. Therefore, considering the nature of Bing as an online platform, it is not possible to hold Bing in a basket. The question is asking about holding Bing, the online platform, in a basket, which is not feasible. Therefore, the answer is: No, you cannot hold Bing in a basket.\\

\hline
The LLM is not able to give a certain answer to the query in the 0-Shot CoT setting as it is unsure about what ``Bing'' refers to. The response provided by the 2-Shot CoT is correct. Also, since $R3$ utilizes the grounded KG facts, it is able to certainly identify what Bing refers to and provide the correct answer.\\
\hline
\end{longtable}

\begin{longtable}{| p{0.95\textwidth} |}
\hline
\textbf{Query:} Did Evander Holyfield compete in an Olympics hosted in the western hemisphere? (Yes)\\
\hline
\textbf{0-Shot CoT:} First, we need to establish if Evander Holyfield competed in the Olympics. Yes, Evander Holyfield \tcg{competed in the 1984 Summer Olympics in Los Angeles, California}, which is in the \tcg{western hemisphere}. Therefore, Evander Holyfield competed in an Olympics hosted in the western hemisphere. Final answer: Yes.
\\
\\
\textbf{2-Shot CoT:} Yes, Evander Holyfield competed in the \tcg{1984 Summer Olympics}, which were hosted in \tcg{Los Angeles, California, United States}. Final answer: Yes.
\\
\\
\textbf{$\mathbf{R^3}$:} The general rule states that Evander Holyfield must have competed in the 1984 Summer Olympics held in the western hemisphere for him to have participated in an Olympics hosted in the western hemisphere. Fact “(Evander Holyfield, participant in, 1984 Summer Olympics)” confirms that Evander Holyfield participated in the \tcg{1984 Summer Olympics}. Fact ``(Evander Holyfield, country of citizenship, United States of America)'' states that Evander Holyfield is a citizen of the United States of America. Since the 1984 Summer Olympics were held in \tcg{Los Angeles, USA}, which is in the \tcg{western hemisphere}, Evander Holyfield did compete in an Olympics hosted in the western hemisphere. Therefore, the answer is: Yes, Evander Holyfield competed in an Olympics hosted in the western hemisphere.\\
\hline
All three methods are able to provide the correct answer to this query.\\
\hline
\end{longtable}

\begin{longtable}{| p{0.95\textwidth} |}
\hline
\textbf{Query:} Can Larry King's ex-wives form a water polo team? (Yes)\\
\hline
\textbf{0-Shot CoT:} First, let's determine how many ex-wives Larry King has. As of my knowledge cutoff date in September 2021, Larry King was reported to have been married \tcg{eight times} to \tcg{seven different women}, with one of them he married twice. If all of Larry King's ex-wives are physically able and willing to participate, they could theoretically form a water polo team, as the sport has no restrictions based on marital status. Therefore, the answer is: Yes.
\\
\textbf{2-Shot CoT:} Larry King has had \tcg{seven ex-wives}. A water polo team consists of \tcg{seven players} in the water and \tcr{one substitute}. Therefore, technically his ex-wives could form a water polo team. Final answer: yes.\\
\textbf{$\mathbf{R^3}$:} According to the general rule, Larry King's ex-wives must possess the skills and abilities required to play water polo in order to form a water polo team. However, the facts provided do not give any information about the skills or abilities of Larry King's ex-wives. Therefore, we cannot determine whether Larry King's ex-wives can form a water polo team or not. The answer is: I don't know.\\
\hline
The response provided by both 0-Shot CoT and 2-Shot CoT are correct, except that 2-Shot CoT has an error about the number of substitute players in water polo which is in fact 6. Due to using an improper commonsense axiom, $R^3$ fails to provide the correct answer. However, instead of providing a hallucinated answer, $R^3$ refrains from answering and states that it does not know the answer. \\
\hline
\end{longtable}

\begin{longtable}{| p{0.95\textwidth} |}
\hline
\textbf{Query:} Would Janet Jackson avoid a dish with ham? (Yes)\\
\hline
\textbf{0-Shot CoT:} First, we need to consider if Janet Jackson follows a dietary restriction that prohibits the consumption of pork or ham. If she follows a kosher diet or has other personal reasons for avoiding ham, then it's possible she would avoid a dish with ham. However, without specific information about her dietary choices, it's difficult to say for certain. Therefore, I don't know.
\\
\\
\textbf{2-Shot CoT:} As Janet Jackson is known to \ul{have followed a vegetarian diet in the past}, it is likely that she would avoid a dish with ham. Final answer: yes.
\\
\\

\textbf{$\mathbf{R^3}$:} The general rule states that Janet Jackson must have a specific dietary restriction or personal preference regarding ham to determine if she would avoid a dish with ham. Fact ``(Janet Jackson, religion or worldview, Sunni Islam)'' tells us that Janet Jackson follows Sunni Islam. In Islam, consuming pork, including ham, is prohibited. Given that Janet Jackson follows Sunni Islam, it is highly likely that she would avoid a dish with ham due to religious dietary restrictions. Therefore, the answer is: Yes, Janet Jackson would avoid a dish with ham.'\\
\hline
\textbf{Explanation:} 0-Shot CoT is unable to provide a specific response. Although the final response provided by 2-shot CoT is correct, it is based on an incorrect reason. The response provided by $R^3$ is correct and based on the correct reason that is based on a fact extracted from the KG.\\
\hline
\end{longtable}

\begin{longtable}{| p{0.95\textwidth} |}
\hline
\textbf{Query:} Would the author of Little Women have remembered ratification of the 13th Amendment? (Yes) \\
\hline
\textbf{0-Shot CoT:} First, we need to establish when ``Little Women'' was published and whether the author, Louisa May Alcott, was alive at the time of the ratification of the 13th Amendment. \ul{``Little Women'' was published} \tcr{in 1868}, and the 13th Amendment was ratified by the states on December 6, \tcg{1865}. Given the timeframe, it is possible that the author would have remembered the ratification of the 13th Amendment. Final answer: Yes.\\
\\
\textbf{2-Shot CoT:}   Louisa May Alcott, the author of Little Women, \tcg{was alive and actively writing} during the time of the American Civil War and the subsequent ratification of the 13th Amendment, which abolished slavery. Therefore, it is likely that she would have remembered the ratification of the 13th Amendment. Final Answer: Yes.\\
\\
\textbf{$\mathbf{R^3}$:} Louisa May Alcott was born in \tcg{1832} and died in \tcg{1888}. The Thirteenth Amendment to the United States Constitution was ratified in \tcg{1865}. Since Louisa May Alcott was alive and cognizant during the time of the ratification of the 13th Amendment, it is plausible that she would have remembered it. Therefore, the answer is: Yes, the author of Little Women, Louisa May Alcott, would have remembered the ratification of the 13th Amendment.\\
\hline
Although all three methods provide the correct final answer to the query, the response provided by $R^3$ is more favorable. The 0-Shot CoT's response contains a hallucination about the publication date of the ``Little Women'' book. In the 2-Shot CoT's response, the exact dates on which the book was published and the time Louisa May Alcott lived are not mentioned, therefore the final response is not verifiable. However, $R^3$ grounds the required facts on the KG and correctly identifies and mentions the dates required for finding the answer which also allows for verification of the final answer.\\
\hline
\end{longtable}
\clearpage
\paragraph{Claim Verification: Long-Tail}
\begin{longtable}{| p{0.95\textwidth} |}
\hline
\textbf{Query:} Melahat Pars is very likely to be familiar with Doner kebab.\\
\hline
\textbf{0-Shot CoT:} Melahat Pars is likely familiar with Doner kebab, as it is a \tcg{popular Turkish dish} and Melahat Pars is a Turkish \tcr{chef} known for her expertise in Turkish cuisine. Final answer: Yes.
\\
\\
\textbf{2-Shot CoT:} As Melahat Pars is a common Turkish name, and Döner kebab is a popular \tcg{Turkish dish}, it's likely that Melahat Pars is familiar with Döner kebab. Final Answer: Yes.
\\
\\
\textbf{$\mathbf{R^3}$:} The general rule states that Melahat Pars, being a Turkish singer and having lived in Turkey, is likely to be familiar with Doner kebab, which is a part of Turkish cuisine. Fact “(Melahat Pars, is, Turkish singer)” tells us that Melahat Pars is a \tcg{Turkish singer} and Fact “(Melahat Pars, country of citizenship, Turkey)” tells us that she is a \tcg{citizen of Turkey}. Fact “(doner kebab, cuisine, Turkish cuisine)” confirms that Doner kebab is indeed a part of \tcg{Turkish cuisine}. Therefore, based on the general rule and the facts, it is very likely that Melahat Pars is familiar with Doner kebab. Therefore, the answer is: Yes, Melahat Pars is very likely to be familiar with Doner kebab.\\
\hline
The 0-Shot CoT response contains a factual hallucination about Melahat Pars' occupation, although the final answer is correct. The 2-Shot CoT response is based on the judgment about Melahat Pars' name, therefore reflecting the fact that the LLM is unfamiliar with this entity. By identifying and utilizing the relevant KG facts, $R^3$ is able to provide a correct and grounded response to the query.\\
\hline
\end{longtable}

\begin{longtable}{| p{0.95\textwidth} |}
\hline
\textbf{Query:} Some people study books about Xuanxue.\\
\hline
\textbf{0-Shot CoT:} 1. What is Xuanxue? 2. Who are "some people"? 3. Why do "some people" study books about Xuanxue? Final answer: Yes.
\\ \\
\textbf{2-Shot CoT:} Xuanxue, also known as \tcg{"Neo-Daoism"}, is a \tcg{philosophical movement in ancient China}. It is \tcg{still studied} and researched by scholars and enthusiasts. Therefore, it is accurate to say that some people study books about Xuanxue. Final answer: Yes.
\\
\\
\textbf{$\mathbf{R^3}$:} The general rule states that some people study books about Xuanxue because it is a philosophical movement in post-classical Chinese history that combines Daoist and Confucian beliefs. Facts \tcg{“(Xuanxue, is, post-classical Chinese philosophy bringing together Daoist and Confucian beliefs)”, “(Xuanxue, instance of, philosophical movement)”, “(Xuanxue, has part(s), Xuanxue in Wei and Jin dynasties)”, and “(Xuanxue, inception, 0300)”} provide information about Xuanxue being a post-classical Chinese philosophy that combines Daoist and Confucian beliefs. Based on this information, it is reasonable to assume that there are books about Xuanxue available for study. Therefore, the answer is: Yes, some people study books about Xuanxue.\\
\hline
The answer provided by 0-Shot CoT lacks a spacific and correct reasoning procedure, although the final answer is correct. The answers provided by both 2-Shot CoT and $R^3$ are correct and based on correct facts and reasoning processes.\\
\hline
\end{longtable}

\clearpage

\begin{longtable}{| p{0.95\textwidth} |}
\hline
\textbf{Query:} One can see the Alaviyan Dome and the Dome of Soltaniyeh at the same time. (No)\\ \\
\hline
\textbf{0-Shot CoT:} The Alaviyan Dome is located in \tcr{Shiraz, Iran}, while the Dome of Soltaniyeh is situated in \tcg{Soltaniyeh, Zanjan, Iran}, and they are approximately 500 kilometers apart. Considering the distance and the curvature of the Earth, it is not possible to see both domes at the same time with the naked eye.\\
\\
\textbf{2-Shot CoT:} The Alaviyan Dome and the Dome of Soltaniyeh are located in \tcg{different cities} in Iran, and are quite far apart from each other. It is not possible to see both domes at the same time due to the distance between them. Final Answer: No.\\
\\
\textbf{$\mathbf{R^3}$:} The sentence claims that one can see the Alaviyan Dome and the Dome of Soltaniyeh at the same time. According to the general rule, the Alaviyan Dome and the Dome of Soltaniyeh must be located in the same place to be seen at the same time. Fact “(Alaviyan Dome, location, Hamadan)” tells us that the Alaviyan Dome is located \tcg{in Hamadan}, Iran. Fact “(Dome of Soltaniyeh, location, Soltaniyeh)” tells us that the Dome of Soltaniyeh is also located in \tcg{Soltaniyeh, Iran}. Since Hamadan and Soltaniyeh are different locations, it is not possible to see the Alaviyan Dome and the Dome of Soltaniyeh at the same time. Therefore, the answer is: No, one cannot see the Alaviyan Dome and the Dome of Soltaniyeh at the same time because they are located in different places.\\
\\
\hline
The the final answer provided by 0-Shot CoT is correct, it contains hallucination about the location of the Alaviyan Dome. Although 2-Shot CoT correctly mentions that the two domes are located in different cities, it does not refer to their particular locations. However, $R^3$'s response is grounded on the KG facts about the specific cities that the domes are located in and the surfaced commonsense axiom, therefore being more specific and easily verifiable.\\
\\
\hline
\end{longtable}

\begin{longtable}{| p{0.95\textwidth} |}
\hline
\textbf{Query:} Mina Ahadi is likely to celebrate Lent. (No)\\
\hline
\textbf{0-Shot CoT:} Lent is a \tcg{Christian observance}, and Mina Ahadi is known for her work as an activist in the areas of secularism and human rights, particularly focusing on the \tcg{plight of atheists} and apostates in Islamic countries. Given her advocacy work, it is unlikely that she celebrates Lent. Final Answer: No.\\
\\
\textbf{2-Shot CoT:} Mina Ahadi is a human rights activist known for her advocacy for secularism and \tcg{against religious authoritarianism}. Given her public stance, it is unlikely that she would celebrate Lent, which is a Christian religious observance. Final answer: No.\\
\\
\textbf{$\mathbf{R^3}$:} The general rule states that Mina Ahadi, being an atheist, is unlikely to celebrate Lent. Fact “(Mina Ahadi, religion or worldview, atheism)” confirms that Mina Ahadi \tcg{identifies as an atheist}. Lent is a Christian observance and is typically celebrated by Christians. Since Mina Ahadi is an atheist and does not follow any religious worldview, it is unlikely that she would celebrate Lent. Therefore, the answer is: No, Mina Ahadi is unlikely to celebrate Lent.\\
\hline
All three methods provide correct answers to this claim.\\
\hline
\end{longtable}
\clearpage

\begin{longtable}{| p{0.95\textwidth} |}
\hline
\textbf{Query:} Árpád Klatsmányi was likely to be aware of the second law of thermodynamics. (Yes)\\
\hline
\textbf{0-Shot CoT:} Árpád Klatsmányi was a Hungarian \tcr{linguist and writer}, not a physicist or scholar in the field of thermodynamics. Without any evidence that he had a background in the study of thermodynamics, it is unlikely that he was aware of the second law of thermodynamics. Final answer: No.\\
\\
\textbf{2-Shot CoT:} Árpád Klatsmányi lived in the \tcr{19th and 20th centuries}, and the second law of thermodynamics was well established by that time. It is likely that he, as a physicist, would have been aware of this scientific principle. Therefore, Árpád Klatsmányi was likely to be aware of the second law of thermodynamics. Final Answer: Yes.\\
\\
\textbf{$\mathbf{R^3}$:} The general rule states that Árpád Klatsmányi must have specific knowledge or education in the field of thermodynamics to be likely to be aware of the second law of thermodynamics. Fact “(Árpád Klatsmányi, occupation, mechanical engineer)” tells us that Árpád Klatsmányi is a \tcg{mechanical engineer}, and Fact “(Árpád Klatsmányi, educated at, Budapest University of Technology and Economics; end time, 1947)” tells us that he was educated at \tcg{Budapest University of Technology and Economics}. Although we don't have explicit information about his knowledge in thermodynamics, it is plausible to assume that as a mechanical engineer, he would have been exposed to the principles of thermodynamics during his education. Additionally, the second law of thermodynamics is a fundamental concept in thermodynamics, and it is likely that someone with a background in mechanical engineering would be aware of it. Therefore, the answer is: Yes, Árpád Klatsmányi is likely to be aware of the second law of thermodynamics.\\
\\
\hline
The 0-Shot CoT's response is incorrect and contains hallucination about Árpád Klatsmányi's occupation. Although the 2-Shot CoT provies a correct response, it contains hallucination about the time Árpád Klatsmányi lived in. $R^3$ provides the correct response by utilizing the KG facts about Árpád Klatsmányi's occupation and education as well as the surfaced commonsense axiom about his familiarity with the second law of thermodynamics.\\
\hline
\end{longtable}

\begin{longtable}{| p{0.95\textwidth} |}
\hline
\textbf{Query:} María Subíes Forcada's home country was among the allied powers. (No)\\
\hline
\textbf{0-Shot CoT:} First, let's determine María Subíes Forcada's home country. María Subíes Forcada was a \tcg{Spanish} \tcr{politician}. During World War II, Spain was officially \tcg{neutral} and did not formally join either the Axis or Allied powers. Therefore, María Subíes Forcada's home country was not among the Allied powers. Final Answer: No.\\
\\
\textbf{2-Shot CoT:} María Subíes Forcada was from \tcg{Spain}, which was a \tcr{member of the Axis powers} during World War II, and did not join the Allied powers until after the war. Final answer: No.\\
\\
\textbf{$\mathbf{R^3}$:} General rule “María Subíes Forcada's home country must be among the Allied Powers during World War II to fulfill the claim.” tells us that María Subíes Forcada's home country must be among the Allied Powers during World War II. Fact “(María Subíes Forcada, country of citizenship, Spain)” tells us that María Subíes Forcada's country of citizenship is \tcg{Spain}. However, the facts do not provide explicit information about Spain's involvement in World War II. We can make plausible assumptions based on general knowledge. Spain was \tcg{not among the Allied Powers} during World War II, but it was a \tcg{neutral country}. Therefore, the answer is: No, María Subíes Forcada's home country was not among the Allied Powers during World War II.\\
\hline
The 0-Shot CoT response contains a hallucination about María Subíes Forcada's occupation. The 2-Shot CoT response contains a hallucination about Spain's involvement in the axis powers during World War II. However, $R^3$'s response does not contain any hallucinations and grounds the information about María Subíes Forcada's nationality on the KG. Furthermore, it does not hallucinate about the side of Spain during World War II.\\
\hline
\end{longtable}

\begin{longtable}{| p{0.95\textwidth} |}
\hline
\textbf{Query:} Noelia Vargas must sleep to survive. (Yes)\\
\hline
\textbf{0-Shot CoT:} No, Noelia Vargas \tcr{does not need} to sleep to survive.\\
\\
\textbf{2-Shot CoT:} Noelia Vargas, like \tcg{all humans}, requires sleep for survival. Final Answer: Yes.\\
\\
\textbf{$\mathbf{R^3}$:} The general rule states that all humans, including Noelia Vargas, must sleep to survive. Fact ``(Noelia Vargas, instance of, human)'' tells us that Noelia Vargas is \tcg{a human}. Therefore, based on the general rule, we can conclude that Noelia Vargas must sleep to survive. Therefore, the answer is: Yes, Noelia Vargas must sleep to survive.\\
\hline
The 0-Shot CoT response is incorrect and includes a hallucination about Noelia Vargas. The 2-Shot CoT and $R^3$ are both able to give the correct answer.\\
\hline
\end{longtable}


\begin{longtable}{| p{0.95\textwidth} |}
\hline
\textbf{Query:} When people wanted to address Ulrik Anton Motzfeldt in a court, they said, Your Honor. (Yes)\\
\hline
\textbf{0-Shot CoT:} First, it's important to establish who Ulrik Anton Motzfeldt is. Ulrik Anton Motzfeldt was a \tcg{Norwegian politician and judge} who lived in the 19th century. During his time, the term ``Your Honor'' was commonly used to address judges in court as a sign of respect. Therefore, it is likely that when people wanted to address Ulrik Anton Motzfeldt in a court, they said "Your Honor." Final Answer: Yes.\\
\\
\textbf{2-Shot CoT:} Ulrik Anton Motzfeldt was a \tcr{Norwegian theologian and politician}, and the phrase ``Your Honor'' is typically used in reference to judges, not theologians or politicians. Therefore, it is unlikely that people would address Ulrik Anton Motzfeldt as ``Your Honor'' in court. Final answer: No.\\
\\
\textbf{$\mathbf{R^3}$:} The general rule ``When addressing Ulrik Anton Motzfeldt in a court, people say ``Your Honor'' because he is a judge.'' states that when addressing Ulrik Anton Motzfeldt in a court, people say ``Your Honor'' because he is a judge. Fact ``(Ulrik Anton Motzfeldt, position held, Justice in the Supreme Court of Norway)'' confirms that Ulrik Anton Motzfeldt \tcg{held the position of Justice in the Supreme Court of Norway}, which indicates that he is indeed \tcg{a judge}. Therefore, the answer is: Yes, when people wanted to address Ulrik Anton Motzfeldt in a court, they said "Your Honor".\\
\hline
In the 2-Shot CoT response, the LLM has neglected the fact that Ulrik Anton Motzfeldt was a judge, but in both 0-Shot CoT and $R^3$ responses, this fact is considered which results in the correct answer.\\
\hline
\end{longtable}

\clearpage

\begin{longtable}{| p{0.95\textwidth} |}
\hline
\textbf{Query:} Jamshid Amouzegar's life began in the third decade of the 20th century. (Yes)\\
\hline
\textbf{0-Shot CoT:} To answer the question, we need to find out when the third decade of the 20th century was. The third decade of a century refers to the 1920s. Therefore, if Jamshid Amouzegar's life began in the third decade of the 20th century, it means he was born in the 1920s.\\
\\
\textbf{2-Shot CoT:} To determine when Jamshid Amouzegar's life began, let's consider the third decade of the 20th century. The third decade spans from 1920 to 1929. If Jamshid Amouzegar's life began in this time frame, then the statement is true. Final answer: I don't know.\\
\\
\textbf{$\mathbf{R^3}$:} According to the general rule, if a person's birth year falls within the specified decade, then their life began in that decade. Fact ``(Jamshid Amouzegar, date of birth, 1923/06/25)'' tells us that Jamshid Amouzegar \tcg{was born in 1923}, which is within the third decade of the 20th century (1921-1930). Therefore, we can conclude that Jamshid Amouzegar's life began in the third decade of the 20th century. Therefore, the answer is: Yes, Jamshid Amouzegar's life began in the third decade of the 20th century.\\
\\
\hline
Since the LLM is unfamiliar with Jamshid Amouzegar, it cannot provide a certain answer in 0-Shot and 2-Shot settings, but since $R^3$ responds based on grounded KG facts, it correctly answers the query.\\
\hline
\end{longtable}

\begin{longtable}{| p{0.95\textwidth} |}
\hline
\textbf{Query:} Chiara Guerra earns a living by portraying different characters in front of a camera. (Yes)\\
\hline
\textbf{0-Shot CoT:} First, let's identify the key elements of the question:
- Chiara Guerra\\
 - Earns a living\\
 - Portraying different characters\\
 - In front of a camera\\
\ul{Based on the information provided}, it suggests that Chiara Guerra's profession involves portraying different characters in front of a camera to earn a living. Final answer: Yes.
\\
\textbf{2-Shot CoT:} I don't have specific information about Chiara Guerra and her occupation. Therefore, I don't know.
\\
\textbf{$\mathbf{R^3}$:} The general rule states that Chiara Guerra must be an actor to earn a living by portraying different characters in front of a camera. Fact ``(Chiara Guerra, occupation, actor)'', ``(Chiara Guerra, is, Italian actress)'', ``(Chiara Guerra, instance of, human)'', and ``(Chiara Guerra, place of birth, Camposampiero)'' all indicate that Chiara Guerra is an \tcg{Italian actress}. Therefore, based on the facts and the general rule, we can conclude that Chiara Guerra earns a living by portraying different characters in front of a camera. Therefore, the answer is: Yes, Chiara Guerra earns a living by portraying different characters in front of a camera.\\
\\
\hline
The response proivided by 0-Shot CoT is completely incorrect as it considers pieces of the claim as reliable information. Since the LLM is unfamiliar with Chiara Guerra, it does not provide a decisive answer in the 2-Shot CoT response. However, the answer provided by $R^3$ is correct and grounded on the KG facts.\\
\hline
\end{longtable}
\twocolumn
\clearpage

\section{LLM Usage in $R^3$}

Several components of the $R^3$ framework make use of an LLM. In this section, we provide explanations about the way that LLM is used in each module and provide the prompts that we used for each LLM-based module. Since prompts for the claim verification and question answering tasks are similar, we provide question answering prompts here, and also release all prompts for the claim verification as well as preference reasoning with our code and data.

\paragraph{Obtaining Relevant Sub-graph}
A key motivation of the KGQA methodologies such as $R^3$ is being able to answer queries about recent and obscure entities. However, existing pre-trained entity extractors are limited to the more famous entities that they were exposed to during their training. Therefore, they may fail to extract recent entities that were not included in the KG at the time of their training or obscure and long-tail entities. To overcome this challenge, as explained in Section \ref{sec:r3}, $R^3$ uses both an off-the-shelf entity extractor and an LLM-based entity extractor and unites the sets of entities both methods return and uses the resulting set to extract their relevant subgraphs. In the ablation study section, we provide an analysis on the role of each entity extractor and provide a discussion on their necessity in $R^3$'s proper performance.

The prompt used in the LLM-based entity extractor is as follows:

\begin{lstlisting}[style=promptstyle]
You are a helpful assistant helping in finding the answer to a question. The found answer has to be based on Wikidata Knowledge Graph triples obtained about entities. Given a question and a helpful fact, identify the least number of entities for which we need to obtain information to be able to solve the question.
  You must only mention the entities and nothing else.
  Write the entities in the following format:
  Selected entity/entities:
  entity1
  entity2...
  [Few-shot Examples]
\end{lstlisting}

\paragraph{Surfacing the Commonsense Axiom}
The commonsense axioms that guide each branch of the tree-structured search in the $R^3$ framework are also surfaced from the LLM. These axioms are critically important in successfully answering the queries. The prompt used for this task is therefore carefully designed to explicitly mention the required desiderata of a useful commonsense axiom. The prompt used for this module is:

\begin{lstlisting}[style=promptstyle]
Task: You are a helpful assistant trying to give us some guidance about answering a question. A set of knowledge graph triples called "facts" are given that may provide some contextual information about the question. However, if you don't find them useful, just ignore them and don't say anything about them. We may later look for additional facts to answer the question. Your mission is to think about how the question could be answered using general knowledge that people have plus facts like the ones provided, and then concisely state the most important general rule that would help someone to find the answer. But, you must not directly answer the question and you must not judge whether the question is answerable or not. Focus on what general information can help in giving a yes/no answer to the question.
Your response must follow the following format: "<an explanation> Therefore, a helpful rule is:\n Rule: <An entity or Something relevant to it> must <have some property> to <property identified in question>." Try your best to use your general knowledge. Be smart. Don't ask or state conditions on obvious information that most average humans would know. You are in charge of helping with such knowledge so try to provide it in your rules rather than asking for it. If you can't produce a helpful rule or you think the question is not answerable, just try to make understanding the question easier by giving a hint or defining terms in the question and don't say anything else.
[Few-shot Examples]
\end{lstlisting}

\paragraph{Sub-graph Pruning}

After surfacing the commonsense axiom, relevant candidate facts from the sub-graph that can be used to ground the answer on them are obtained by using both an LLM-based module and also semantic similarity between the embedding vectors. Prompts used for the LLM-based sub-graph pruning module is as follows:

\begin{lstlisting}[style=promptstyle]
Task: You are a helpful assistant that is trying to help us answer a question. Given the question, a general rule that will help us answer the question, and a list of knowledge graph triples which we call them facts. Consider the facts and think about their relation to the question and general rule and try to extract the facts that may help answering the question. The facts may be insufficient to answer the question, but try your best to extract the relevant facts.
Your response must follow this format:
<an explanation> Therefore, the relevant facts are: <list of relevant facts>
Just copy the selected facts and don't generate facts on your own or adjust the facts in any way. Try your best to select the relevant facts. If there are no relevant facts, just output "None".
[Few-shot Examples]
\end{lstlisting}

\paragraph{Fact-Grounded Answer Selection}
In the light of the retrieved relevant facts, the LLM tries to select the answer. In the prompt used for this module, we aim to clarify for the LLM to try to answer the question if the provided facts are sufficient, and otherwise respond with ``I don't know''. The prompt used for this module is as follows:

\begin{lstlisting}[style=promptstyle]
Task: You are a helpful assistant that is trying to help us answer a question. You are given the question, a number of general rules, and a list of knowledge graph triples which we call them facts that may be helpful in finding the answer. First, go over the facts and general rules one by one. Try to think of how each fact may help you answer the question. Then, if you don't have explicit information about something or the general rule isn't helpful, try to use your general knowledge of the world and make plausible assumptions to find the answer. Be smart. Don't ask for obvious information that most average humans would know.
Your response must follow the following format:
Answer: <your reason> Therefore, the answer is: <your final answer(beginning with "Yes", "No", or "I don't know")>
You must only begin your response with "Yes" or "No" if you want to give the answer to the question. Try your best to use facts, general rules, and plausible assumptions to give the answer. If using the current set of general rules and facts is not enough to answer the question even with plausible assumptions, in the beginning of your answer, you must only say "I don't know".
[Few-shot Examples]
\end{lstlisting}

\paragraph{Missing Evidence Identification}
In case the LLM determines the existing facts to be insufficient, we need to identify what missing evidence is required. This performance is obtained in two steps. First, the LLM is asked to identify what missing information is required, for which the following prompt is used:
\begin{lstlisting}[style=promptstyle]
Task: You are a helpful assistant trying to help in finding the required information to answer a given question. A the set of general rules and a list of knowledge graph triples, which we name facts, are already provided. Based on these, an answer was propsed, but it was not identified as being correct and certain. You are asked to identify what other facts are required to give a certain answer to the question. The facts you ask for will be obtained from a knowledge graph. So, try to extract the name of entity or entities about which we should obtain facts and mention it in your answer. For example, if knowing about Bill Clinton's daughter's religion is necessary, and among the already provided facts you see ('Bill Clinton', 'child', 'Chelsea Clinton'), you should respond "we need to know Chelsea Clinton's religion".
Finally If the provided facts and general rules are already sufficient to give a certain answer to the question, your response should only be: "nothing".

[Few-shot Examples]
\end{lstlisting}
Next, we ask the LLM to identify the next entity for which we need to obtain the relevant sub-graph to continue the search branch. For this step, the following prompt is used:

\begin{lstlisting}[style=promptstyle]
Task: Considering the provided information need that is needed to answer the question and a set of relevant facts, identify the name of the Wikidata entity that facts about it will be helpful in fulfilling the information need. Try to extract the entity name from the relevant facts. For example, if the information need states that we need to know about Bill Clinton's daughter, use the fact ('Bill Clinton', 'child', 'Chelsea Clinton') and select the entity name Chelsea Clinton. Remember that the entity name you pick must be different from all Previously chosen entities.
[Few-shot Examples]
\end{lstlisting}

\end{document}